\documentclass[preprint,12pt,3p]{elsarticle}
 
\usepackage{amssymb}
\usepackage{url}
\usepackage{caption}
\usepackage{subcaption}
\usepackage{color}
\usepackage{multirow}
\usepackage{amsmath}

\journal{Computer Speech and Language}

\begin{document}

\begin{frontmatter}

\title{Automatic Quality Estimation for ASR System Combination}

 \author[label1,label2]{Shahab Jalalvand \corref{cor1}}
 \address[label1]{FBK-Fondazione Bruno Kessler, Trento, Italy}
 \address[label2]{University of Trento, Italy}

 \cortext[cor1]{\textcolor{red}{Please cite this paper as: S. Jalalvand  et al., Automatic Quality Estimation for ASR System Combination, Computer Speech \& Language (2017), http://dx.doi.org/10.1016/j.csl.2017.06.003} }

 \ead{jalalvand@fbk.eu}

 \author[label1]{Matteo Negri \corref{cor1}}
 \ead{negri@fbk.eu}

 \author[label1]{Daniele Falavigna}
 \ead{falavi@fbk.eu}

\author[label1]{Marco Matassoni}
 \ead{matasso@fbk.eu}

\author[label1]{Marco Turchi}
 \ead{turchi@fbk.eu}

\begin{abstract}
Recognizer Output Voting Error Reduction (ROVER) has been widely used for system combination in automatic speech recognition (ASR). In order to select the most appropriate words to insert at each position in the output transcriptions, some ROVER extensions rely on  critical information such as confidence scores and other ASR decoder features. This information, which is not always available, highly depends on the decoding process and sometimes tends to overestimate the real quality of the recognized words. In this paper we propose a novel variant of ROVER that takes advantage of ASR quality estimation (QE) for ranking the transcriptions at ``segment level" instead of: \textit{i)} relying on confidence scores, or \textit{ii)} feeding ROVER with randomly ordered hypotheses. We first introduce an effective set of features to compensate for the absence of ASR decoder information. Then, we apply QE techniques to perform accurate hypothesis ranking at segment-level before starting the fusion process. The evaluation is carried out on two different tasks, in which we respectively combine  hypotheses coming from independent ASR  systems and multi-microphone recordings. 
In both tasks, it is assumed that the ASR decoder information is not available.
The proposed approach significantly outperforms standard ROVER and it is competitive with two strong oracles that exploit prior knowledge about the real quality of the hypotheses to be combined. Compared to standard ROVER, the absolute WER improvements in the two evaluation scenarios range from 0.5\% to 7.3\%.
\end{abstract}


\end{frontmatter}

\section{Introduction}
\label{intro}
The application of 
ASR 
systems in our daily life is steadily increasing. Voice search engines, voice question answering, broadcast news transcriptions, video/TV programs subtitling, meeting transcriptions  and spoken dialog systems are just some of the many applications involving ASR technology. In such applications, the quality of  transcriptions and the availability of fast and accurate automatic evaluation methods represent two crucial  interconnected needs. Automatic ASR evaluation, indeed, does not only represent a way to assess system performance \textit{a posteriori}, but also a way to improve it. 
For instance, automatic assessment methods can be used to select audio data for  unsupervised training 
{\cite{Lamel2001, fala2016}} and active learning of  acoustic models \cite{riccardi2005active,falavignaSC2006} or, as in the case of this paper, to combine multiple transcription hypotheses into a single and more accurate one.

In order to synthetically obtain more accurate transcriptions, ASR systems’ diversity and complementarity have been exploited in different ways \cite{audhkhasi2014theoretical}. The combination of multiple hypotheses coming from independent sources  usually leads to significant improvement compared to the output of each individual system. ROVER,
the most popular ASR system combination approach,  performs  hypothesis fusion by first building a word confusion network (CN) from the \textit{1}-best hypotheses of the ASR systems entering the combination and then by selecting the best word in each CN bin 
via majority voting \cite{fiscus1997post}. 
{Word confidence scores, when available, can be exploited to inform the process by setting up a weighted majority voting scheme. Although several enhancements of this general strategy have been successfully proposed, ROVER-based hypothesis combination methods still suffer from some limitations that this paper aims to overcome.}

{The first one 
depends on how they are generally implemented: the hypothesis combination process considers the first input candidate as a ``skeleton'' to align the other hypotheses in a greedy manner. For this reason, depending on 
the  order in which the hypotheses are considered when feeding the algorithm, the resulting combination can show large variations in quality. 
{This raises the need of reliable techniques for ranking the input hypotheses before starting the fusion process.}}

{The second  limitation relates to the granularity of the input 
fed into ROVER: in the standard setting, transcription hypotheses correspond to entire audio recordings, whose duration can span up to hours. 
However, when 
manipulating long utterance transcriptions, the skeleton used to initialise the process can feature significant local variations in terms of quality. 
As a consequence, it may happen that 
the worst 
hypothesis  for an entire audio recording (\textit{i.e.} globally) is the best one for one or more passages (\textit{i.e.} locally). 
{This raises the need of strategies for operating at higher levels of granularity 
(\textit{e.g.} segments spanning over few seconds) 
in order to take full advantage of local quality differences between the input hypotheses.}}

{The third limitation concerns the applicability of ROVER-based hypothesis combination techniques: results' quality significantly increases when 
the availability of confidence scores makes it possible to set up
weighted majority voting schemes. However, having access to the ASR systems' inner workings 
is a  rigid constraint that limits the applicability of 
hypothesis fusion to scenarios in which the input transcriptions are produced by known ASR tools. Often, however, the hypotheses come from ``black-box'' systems, without additional scores.\footnote{For instance, this is the case of the steadily increasing volume of Youtube videos 
for which automatic captions are provided by  black-box ASR technology. In 2012,
more than 157 million  videos were already accessible with auto-captions in 10 languages  (source: http://goo.gl/5Wlkjl).}
 {This raises the need of  methods that are independent from confidence information, but still capable to achieve good results with simple frequency-based voting.}}

Finally, it is worth noting that the confidence scores proposed by previous ASR literature 
\cite{Evermann2000,wessel2001confidence}, even when  applicable, only indicate how confident the system is about its own  output. This can be a biased perspective (influenced by individual decoder features), producing scores that are not comparable across different systems. 
{External and system-independent  measures of goodness  would represent a more reliable alternative when comparable and objective ASR quality judgements are required}.

To cope with these issues, in \cite{negri2014quality} we proposed a reference-free and confidence-independent ASR quality estimation (QE) method, in which a supervised regression model is used to predict  the word error rate (WER) of  automatically transcribed audio recordings.
Experimental results 
in different evaluation settings
showed that our QE predictions can closely approximate the true WER scores calculated over reference transcripts. Building on these positive results, in \cite{sjalalvand-EtAl:2015:ACL} we applied ASR QE to inform system combination with ROVER, outlining the framework that this paper aims to extend and  refine. Our study focused on comparing the standard {\em system-level} ROVER with an alternative {\em segment-level} strategy
that uses ASR QE to rank the input hypotheses before starting the fusion process. Our method was applied to combine the transcriptions of English TED talks produced by the eight participants in the IWSLT2013 evaluation campaign.\footnote{The International Workshop on Spoken Language Translation (IWSLT -- \url{http://workshop2013.iwslt.org/}) is a yearly workshop associated with an open evaluation campaign on spoken language translation.} Its results outperformed the standard ROVER (based on averaging the results over a large number of random rankings of ASR outputs) and significantly approached oracle upper-bounds. A preliminary discussion about the  influence of hypothesis diversity on the fusion results was also proposed (an issue theoretically studied also in \cite{audhkhasi2014theoretical}). However, a deeper investigation on how to better exploit the  trade-off between 
the number of combined 
hypotheses and their diversity was left for future works.

To further assess the effectiveness and the general applicability  of the proposed QE-informed ROVER, this paper extends our experiments to a new and  different scenario involving multiple distant microphone (MDM) speech recognition in noisy environments. 
The new experiments are carried out with the data delivered for the $3^{rd}$ CHiME challenge\footnote{The CHiME Speech Separation and Recognition Challenges are international initiatives proposed in 2011, 2013, 2015 and 2016 --  \url{http://spandh.dcs.shef.ac.uk/chime_challenge/}.} \cite{CHiME3_paper}, in which six  microphones, placed on a tablet PC, were used to record sentences of the Wall Street Journal (WSJ) corpus \cite{paul1992design} uttered by different speakers in four environments (bus, cafe, pedestrian area and street junction). In this case, the diversity among the combined ASR hypotheses is caused by different positions of the microphones that acquire the audio signals, while the ASR system that processes the different audio streams remains unchanged.

Although MDM hypothesis combination can be considered as an alternative to the signal enhancement techniques proposed in the  past \cite{wolfel2009distant,kumatani2011channel,mestre2003signal},
in this work we show that QE-informed ROVER yields further improvements, even after including the enhanced channels in the combination. Similar to \cite{sjalalvand-EtAl:2015:ACL}, we  first use ASR QE techniques to predict the quality of each channel transcription (including the enhanced channels) and then we rank them before applying ROVER. To the best of our knowledge, this is the first time that ASR QE is applied for ranking MDM hypotheses.

Building on the positive results presented in \cite{sjalalvand-EtAl:2015:ACL}
and by discussing a new set of experiments, this paper shows that: 

\begin{itemize}
\item ASR QE can support system combination with ROVER  by providing reliable ranks of the  hypotheses produced by different ASR systems.  For the automatic transcriptions of TED talks in the IWSLT dataset, the absolute  WER  reductions
with respect to the best individual system and standard random ROVER are respectively 1.6\% and 0.5\%;

\item ASR QE consistently works well also for hypothesis-level combination in the MDM speech recognition scenario (same ASR system but multiple microphones). On CHiME-3 data, 
The WER reductions with respect to the best enhanced channel and random ROVER are respectively 1.7\% and 7.3\%.
\end{itemize}

Moreover, by extending our analysis to new problems, in this work we overcome two limitations of method originally proposed in \cite{sjalalvand-EtAl:2015:ACL}:

\begin{itemize}
\item Dealing with tied ranks. Often, in the training data, several transcriptions of the same segment have the same WER in spite of their different quality. In such cases, having the same label associated to different features reduces the performance of our ranking algorithms. 
We propose a simple solution to this problem, which allows us to obtain better and coherent results over the two datasets.

\item Finding an optimum level of combination. Usually, combining all the available hypotheses does not  lead to the best solution. ROVER, indeed, benefits not only from the quality of the combined transcriptions, but also from their diversity. By investigating this trade-off, we provide an empirical method to select 
an optimum number of hypotheses to combine for each segment (\textit{i.e.} locally). Our results closely approximate those obtained, on both datasets, by the best combinations at global level.
\end{itemize}

The paper is organized as follows. After an overview of related work in \S \ref{related}, we
shortly introduce how ROVER works in \S \ref{rover}. This allows us to set the stage for formalizing our approach to system combination and describe its  overall workflow in \S \ref{method}. Then, together with numeric evidence emphasizing the potential of  our QE-informed approach,
in \S \ref{qe-rover} we provide more details about the algorithms and the features used. Our experimental setting is fully described in \S \ref{experiment}, followed by a discussion of the results achieved on 
IWSLT
and CHiME-3 data in \S \ref{results}. Finally,  \S \ref{diss} and \S \ref{conclusion} respectively investigate new problems (dealing with tied ranks and automatically finding the optimum level for combination) and conclude the paper with final remarks.

\section{Related work}
\label{related}
Scientific literature related  to the work described in this article spans over several lines of investigation such as ASR quality estimation, ASR system combination, signal enhancement and speech recognition  with multiple distant microphones in noisy environments.

\paragraph{Quality Estimation} Differently from confidence estimation, which has been widely studied over the last two decades 
\cite{Evermann2000, wessel2001confidence},
ASR QE is a rather new task. The problem, at its core, shows a strong parallelism with QE in the machine translation (MT) field, where the goal of bypassing the need of manually-created reference translations has motivated a large body of research. The motivations (cost effective quality prediction at run-time) and the methods (supervised learning, either as regression or multiclass classification) are indeed the same in both fields. For a complete overview of the current approaches to MT QE, we refer the reader to the comprehensive overviews published within the yearly Workshops on Statistical Machine Translation \cite{callisonburch-EtAl:2012:WMT,bojar-etal_WMT:2013,bojar-EtAl:2014:W14-33,bojar-EtAl:2015:WMT} and to the works dealing with quality prediction at word level \cite{Ueffing:2007:WCE:1245134.1245137,Bach2011}, sentence level \cite{Specia09,Turchi2013,Souza2013,Souza2014a} and document level \cite{Soricut2010}.

In the ASR field, confidence estimation strongly relies on knowledge about the inner workings of the decoder that produces the transcriptions. Such knowledge includes acoustic and language model scores and word timing information associated to arcs of word lattices. QE, instead, 
approaches the problem as a system-independent task, in which  this information is not necessarily accessible. In the first work along this direction \cite{negri2014quality}, we explored ASR QE as a supervised regression problem in which the WER of an utterance transcription  has to be automatically predicted. Given a set of 
(\textit{signal}, \textit{transcription}, \textit{WER})
triplets as training instances, different algorithms were evaluated on test data consisting of  unseen 
(\textit{signal}, \textit{transcription})
pairs. All models used an effective set of signal/textual/hybrid features, 
named {\em black-box} features in contrast with the {\em glass-box} ones that 
model
the internal behavior of the decoder. Results showed that 
ASR QE predictions can 
closely approximate the ``true'' WER scores calculated over reference transcripts, outperforming a strong baseline in a variety of test conditions of increasing complexity. Interestingly, when combined with the glass-box features (\textit{i.e.} confidence scores), the predictions are better than those obtained by using confidence scores alone.

In \cite{cdesouza-et-al:2015:Naacl}, ASR QE was explored by focusing on the problem of domain mismatches between training and test data. Indeed, as pointed out by \cite{negri2014quality}, simple supervised learning methods are very sensitive to large variations in the distribution of the instances in the two sets (both at the level of labels and at the level of features). The proposed solution relies on multitask learning to train robust models that exploit 
similarities and differences 
between possibly related tasks, 
transferring knowledge across them.
Results show that the approach is able to take advantage of data coming from such heterogeneous domains and to significantly improve over single-task learning baselines both in regression and in classification. These findings 
indicate 
the reliability of ASR QE in particularly challenging test conditions that are out of the scope of this work. Casting ASR QE for system combination as a multitask learning problem, however, is certainly an aspect that we will explore in the future.

\paragraph{System combination}
Most of the approaches proposed in the past for combining multiple ASR outputs make use of word lattices \cite{xiang2002,lium2013}. 
The underlying idea is to merge the word lattices generated by different ASR systems into a single one,
which is then traversed to search for the best path.
As an alternative, frame-based 
system combination \cite{hoffmeister2006frame} tries to minimize a  cost function called ``time frame word error" (fWER) over a set of word lattices produced by different ASR systems. The method makes it possible to estimate the path exhibiting the minimum Bayes risk, without the necessity of merging edges of single word lattices. Confusion network combination (CNC) \cite{mangu2000,hoffmeister2006frame,evermann2000posterior} is another widely investigated approach, in which confusion networks built from the individual lattices are aligned instead of single best outputs.
{More recent approaches  \citep{yang2016system, wang2015joint}  use log-linear models, 
also 
exploited in machine translation tasks 
\cite{KITatIWSLT16},
in which the acoustic likelihoods coming from  different systems are weighted and summed at frame level. The resulting weighted  likelihoods are decoded to produce the final hypotheses. In this way, the combination is applied at a much finer level than that of whole hypotheses (such as CNC or ROVER),
resulting in significantly better performance in a standard noise-robust speech recognition task \citep{yang2016system}.}
Note that all  combination methods based on the use of word lattices, as well as some extensions of ROVER \cite{zhang2006,hillard2007rover,abida2011crover}, require  to know and have access to the inner structure of ASR decoder.
{In most cases this is reflected into the computation of confidence measures for the recognized words.}
Often, however, ASR systems
(especially those embedded in commercial applications) do not provide this information. Instead, standard ROVER and the QE-informed ROVER proposed in this paper do not necessarily need to know the ASR inner behaviour. 

Another limitation of previous ROVER-based methods is that, assuming a fixed order in the quality of the systems that generated the hypotheses, they do not apply any segment-level ranking, disregarding the possible advantages of operating at a higher granularity level.
{The work reported in \cite{schwenk2000improved} proposes to use the probability given by a LM to solve the ambiguity (see \S \ref{rover}) that arises when different words receive the same vote at each confusion bin in the ROVER word network. Inspired by this work, in \cite{sjalalvand-EtAl:2015:ACL}, we presented the first application} of  ASR QE as a way to inform system combination with ROVER.  Different solutions to rank the input hypotheses at \textit{segment level} were compared against the  standard  ROVER that operates at \textit{system level}. In addition to transferring the QE problem into a machine-learned ranking task, we enriched the feature set 
used in \cite{negri2014quality,cdesouza-et-al:2015:Naacl}
by adding word-based features previously proposed in  \cite{litman2000predicting, pellegrini2010improving}  for word error detection tasks \cite{goldwater2010words, tam2014asr, sjalalvand2015stacked}. The approach, which is summarized in \S \ref{method}, was  tested on a specific scenario (the transcription of the English TED talks from IWSLT) and achieved promising results that are also reported in this paper (\S \ref{subsec:iwsltResults}). Improving the method, addressing  the need for a stopping criterion to avoid entering useless inputs into the ROVER combination and extending the evaluation to new scenarios were left as future works in \cite{sjalalvand-EtAl:2015:ACL} and represent the main extensions discussed in this paper.

\paragraph{Multiple Distant Microphone (MDM) speech recognition}
{The use of multiple microphones in 
ASR  is mostly motivated by the necessity to reduce the effects of environmental reverberation and noise over the speech signal to be recognized in distant-talking scenarios.
To do this, delay-and-sum beamforming (DS) is a widely used technique to combine multiple 
recordings. It consists in summing the input signals 
after having temporarily aligned them in order  to compensate for the different times of arrival of the sound waveforms to the various microphones \cite{wolfel2009distant}. 
This approach  has been successfully applied to the recognition of 
voice commands in a domestic environment
\cite{vacher2012recognition,mowlaee2013iterative} and it was also used in the CHiME challenge \cite{barker2013thepascal}. Another well known approach to generate an ``enhanced'' speech waveform  from multiple recordings is based on   minimum variance distortionless response (MVDR) \cite{Kumatani2012}. 
This solution has been applied as speech enhancement baseline in the CHiME-3 scenario \cite{CHiME3_paper}.
}

{In addition to DS, 
MVDR and other signal-based
combination approaches \cite{mestre2003signal}, the automatic selection of the best channel can be used to perform signal enhancement. The simplest method to do this consists in measuring the signal-to-noise ratio  (SNR) of the recorded signals \cite{Wölfel06multi-sourcefar-distance}, assuming that the channel with the highest SNR will be the easiest to transcribe.
Choosing 
the channel for which the ASR engine has generated the highest confidence score 
can also represent a viable solution.
For instance, in a previous work with data from the aforementioned CHiME-3 challenge, we successfully applied channel selection based on sentence confidence estimation~\cite{jalalvand2015boosted} to select the best microphone.
}

{An alternative to 
signal-level combination and 
best microphone selection
is represented by hypothesis-level combination.
Along this direction, \cite{Wölfel05combining} proposes to use  CNC for lecture transcriptions and \cite{stolcke2011making}  extends it with a  hybrid approach that leverages  beamforming and signal-level diversity to transcribe meetings. In \cite{guerrero2014signal}, inter-microphone agreement is used to build a confusion network from multiple lattices and improve ASR in a domotic application.}

Our QE-informed ROVER solution is also compared against MVDR and DS processing and in addition, 
we consider both MVDR and DS hypotheses as enhanced channels and we
include them in ROVER combination, showing performance improvements with respect
to original channel combination.

\section{ROVER}
\label{rover}

\begin{figure}[t]
    \centering
    \includegraphics[trim=0.2cm 0.0cm 0.1cm 0cm, clip=true, width=0.6\linewidth, height=9cm]{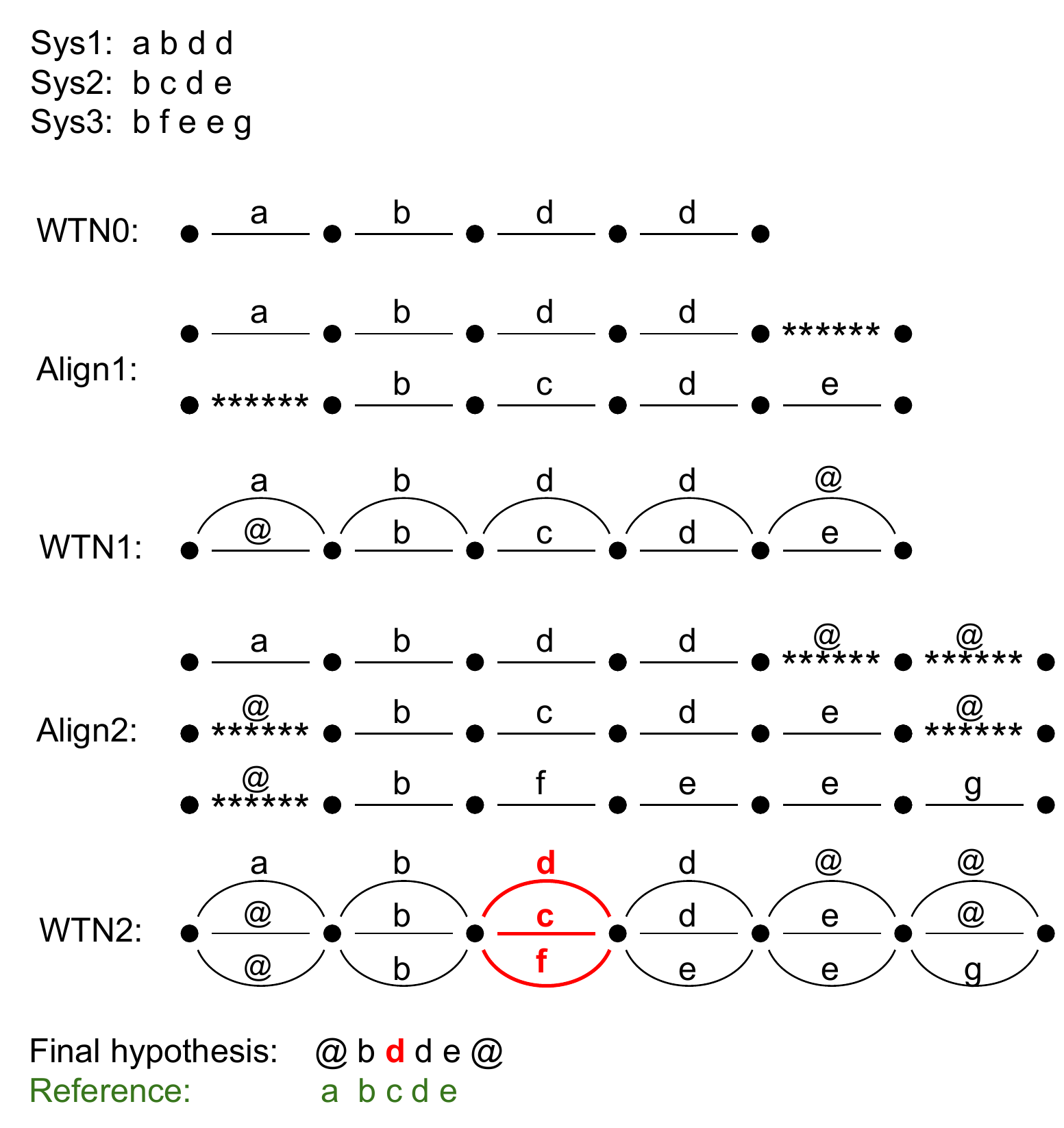}
    \caption{ROVER procedure. $Sys2$ has recognized better than $Sys1$ though it is in the second order. This mistake in the input arrangement, leads to an error in the final hypothesis, while it could be recovered with correct arrangement.}
    \label{fig:roverproc}
\end{figure}

Before discussing our QE-informed approach to hypothesis fusion, in this section we introduce ROVER and the way it is commonly used to combine multiple transcriptions.
Figure~\ref{fig:roverproc} illustrates  an example in which three ASR hypotheses 
(Sys1, Sys2 and Sys3), have to be combined (for the sake of clarity, the words in each hypothesis are represented with letters). 
Given the input transcriptions
 ROVER builds  a word transition network (WTN) by 
exploiting iterative dynamic programming.
Initially, this is done by  selecting the first input hypothesis (Sys1) as a ``skeleton'' to build an initial base WTN ($WTN0$).
Then, the second hypothesis (Sys2) is aligned with $WTN0$  using dynamic programming and 
Levenshtein distance, 
resulting in the  alignment depicted in Figure~\ref{fig:roverproc} ($Align1$), where the asterisks  represent both inserted and deleted words.
Such alignment is used to build a composite WTN ($WTN1$) by using the symbol ``@'' to represent 
insertions and deletions.
The process is 
repeated with the 
next
hypothesis (Sys3),
which results in the final composite $WTN2$\footnote{Note that  to optimally align an hypothesis with a composite WTN a multidimensional dynamic programming would be required, where each dimension is treated as an input hypothesis. Since the implementation of such algorithm is difficult, ROVER \cite{fiscus1997post} uses the approximated solution here summarized.}.

The last WTN, resulting from the 
multiple alignment, can be interpreted as a word confusion network (CN) \citep{mangu2000} with a sequence of word bins.
Starting from this CN, the combined hypothesis is generated by selecting the best word at each bin via majority voting. In case different words receive the same vote (see the red bin in Figure \ref{fig:roverproc}), ROVER gives priority to the word 
in the first position of the bin (\textit{i.e.} ``d'' from Sys1).
This solution for resolving ties raises one of the important problems addressed in this paper. In fact, since the first hypothesis used as skeleton highly influences the voting mechanism, 
the final result will depend on its quality: bad transcriptions can depend from the priority given to a poor-quality skeleton.
To mitigate this problem, the approach discussed in the next sections addresses 
automatic methods for ranking the input hypotheses based on their estimated quality.

Traditionally, ROVER is run by applying
random or arbitrary fixed rankings of the input hypotheses at \textit{system-level}. By system-level ranking we refer to an ordering
that is assigned to the ASR systems included in the combination. This ranking is the same for all the segments in the dataset. 
In our example, if the proposed ranking for the three ASR systems is $Sys1 \preceq Sys2  \preceq Sys3$, then this ranking will be kept unchanged for the entire utterance.
As a result, 
ROVER 
does not exploit
local quality differences in the transcriptions of the utterance. 
In contrast, 
the approach proposed in the next sections
relies on a \textit{segment-level} ranking that dynamically  varies from one segment to another based on their predicted quality. This solution makes it possible to take advantage from the fact that different ASR systems (or microphones) show different performance 
over different portions of the utterance.

\section{General workflow}
\label{method}

\begin{figure}[h]
\centering
\includegraphics[trim=0cm 0cm 0cm 0cm, clip=true,width=\linewidth]{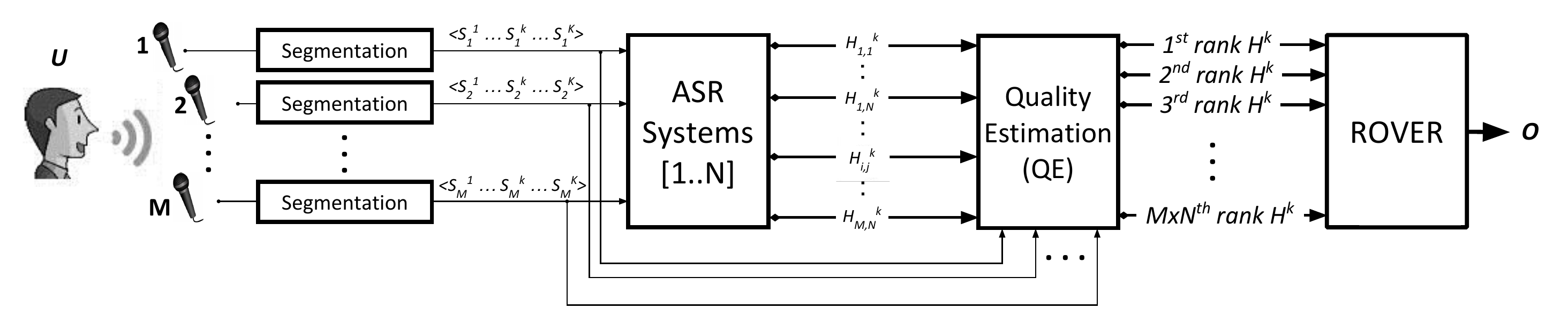}
\caption{The workflow of segment-based QE-informed ROVER. }
\label{arcpic}
\end{figure}

{The workflow of our segment-based QE-informed ROVER is illustrated in Figure~\ref{arcpic}.
As previously mentioned we address two application scenarios: \textit{i)} the combination of transcriptions of a single recording 
coming from $N$ different ASR systems, and \textit{ii)} the combination of  transcriptions of $M$  recordings 
coming from $N$ different ASR systems. Figure~\ref{arcpic} refers to the general case: multiple microphones ($M>1$), multiple ASR systems ($N>1$). }

Audio recordings  are processed by a system that detects  low energy regions (applying specific energy thresholds 
and produce segments of lower duration. Then, each segment is assigned to one of two classes, ``speech'' or ``not speech'', and only the speech segments are passed to the ASR systems 
for transcription.

The automatically detected speech segments $\{S_{1}^{k}\ldots S_{M}^{k}\}$ ($1\leq k\leq K$), resulting from the segmentation of the input utterance $U$, are sent to $N$ ASR systems producing, for each segment $S_j^k$,  $N$ different hypotheses  $\{H_{j,1}^{k}\dots H_{j,N}^{k}\}$. 

The signals and the 
hypotheses
are passed to a QE component.
The QE model is 
trained to predict  the ranking of the hypotheses by using 
features
extracted from the signals 
and the 
transcriptions.
Based on the determined ranking, ROVER combines  the hypotheses of the $k^{th}$ segment $S_j^{k}$ ($1\leq j\leq M$) into $o^{k}$. Finally, the combined  segment-level 
hypotheses $o^k$ 
are concatenated in a single output $O=\{o^{1}+o^{2}..+o^{K}\}$ for the entire audio track.

Note that, in order  to apply this approach, 
the %
same speech segmentation $\{S_{1}^{k}\ldots S_{M}^{k}\}$ must be shared across all microphones and ASR systems. However, this is not usually true because microphones capture voices at different levels of energy and ASR systems make use of different  automatic segmentation algorithms.
A possible solution to address this issue is to extract a common segmentation (derived from an enhanced signal or directly available from a manual segmentation).  

Instead, when 
dealing with multiple ASR systems segmentation, an alignment with a reference segmentation needs to be applied. This is the case of the  experiments with single microphone multiple ASR systems, 
as explained in \S \ref{experiment}.

The general approach depicted in Figure \ref{arcpic} is suitable for a range of conditions in which the hypotheses to be combined come from multiple microphones, multiple ASR systems or any combination of the two. The experimental framework considered in the remainder of this paper focuses on the following two scenarios: 
\begin{enumerate}
\item Single microphone ($M=1$), multiple systems ($N>1$). This is the case of the automatic transcription of TED talks 
from the IWSLT campaign 
(see  \S \ref{subsec:tedd}).
\item Multiple microphones ($M>1$), multiple systems ($N>1$). This is the case of the automatic transcription of  WSJ articles 
from the CHiME-3 campaign 
(see \S \ref{chime_exp}).
\end{enumerate}

The following sections describe how this general workflow has been 
instantiated 
and evaluated in the two scenarios.

\section{Segment-level QE-informed ROVER}
\label{qe-rover}

\subsection{Preliminary analysis} 
\label{pilot}

To investigate the impact  on ROVER results  of considering different rankings and different levels of granularity of the input hypotheses,
we run some pilot experiments on the same data used in  \S \ref{experiment}, namely: TED talks transcriptions (collected with a close talk microphone)  and WSJ read sentences (recorded with distant microphones). In the first case we used the submissions of the IWSLT2013 ASR evaluation campaign \cite{cettolo2013report}, where $8$ teams submitted their ASR results, each containing hypothesized word sequences and related  time boundaries for each  talk. In the second case, we used the data provided for the $3^{rd}$ CHiME challenge \cite{CHiME3_paper}, where each utterance is recorded by $6$ microphones embedded on a tablet PC. One of these microphones, however, is ignored in all our experiments as it is placed on the back of the tablet and mainly captures the background noise of the environment.
The 
CHiME-3
recordings are transcribed using two baseline ASR systems provided by the organizers: 
one employs Gaussian mixtures to compute acoustic probabilities, while the other uses deep neural networks. 
Therefore, in the first task 
({\em IWSLT} henceforth) we combine up to $8$ different hypotheses, while in the second task ({\em CHiME-3}) we combine up to 
$10$ different hypotheses. 
Our goal is to check if, and to what extent, an informed (oracle) ranking at system/segment level can positively contribute to  ROVER results.

\begin{table*} [htbp]
\begin{center}
\begin{small}
\begin{tabular}{|l|p{1cm}p{1cm}p{1cm}p{1cm}||p{1cm}p{1cm}p{1cm}p{1cm}|}
\hline
\multirow{2}{*}{\textbf{Ranking}} & \multicolumn{4}{c||}{IWSLT (best WER=13.5\%)} & \multicolumn{4}{c|}{CHiME-3 (best WER=32.6\%)}   \\ 
& L1 & L3 & L5 & L8 & L1 & L3 & L5 & L10  \\ \hline\hline
SysO  & 13.5 & 12.2 & \textbf{11.8} & 12.1 & 32.6 & 30.7 & \textbf{27.9} & 29.4  \\ 
SegO  & \textbf{8.9} & 10.5 & 11.4 & 11.7 & \textbf{19.5} & 21.4 & 22.4  & 28.4 \\ \hline
InSysO & 27.2 & 19.8 & 15.1 & 13.3 & 40.5 & 37.5 & 34.0 & 29.7\\
InSegO & 33.8 & 22.9 & 17.4 & 13.0 & 56.3 & 51.0 & 42.6 & 30.9\\ \hline
\end{tabular}
\end{small}
\end{center}
\caption{\label{motivation} WER[\%] ($\downarrow$) results of different hypothesis ranking strategies on IWSLT and CHiME-3 test data.}

\end{table*}

Table~\ref{motivation} gives the
oracle scores
at system (SysO) and segment level (SegO) when the best $k$  hypotheses ($Lk$) are used for the combination (for IWSLT $k=1,3,5,8$ while, for CHiME-3, $k=1,3,5,10$).
At both levels of granularity, oracle rankings are obtained by measuring the true WER of the candidate transcriptions.
As it can be seen in the table, hypothesis combination allows for considerable WER reductions compared to the best individual system,  both at system and at segment-level. In particular, on IWSLT, system-level combination results in a WER reduction from 13.5\% (best individual system) to 11.8\% when combining the best 5 systems. Similarly, on CHiME-3, WER decreases from 32.6\% 
(best individual channel\footnote{Henceforth, for CHiME-3 data, the terms ``channel'' and ``microphone'' will be  interchangeably used.}) 
to 27.9\% when the best 5 hypotheses are taken. Note that, in both tasks, system-level combination achieves the best results with only 5 different hypotheses, which is less than all the available ones.\footnote{The fact that the best level of combination is the same on both datasets  is purely incidental. As we will see in  \S \ref{subsec:optimumcomb}, the definition of the optimum level of combination for a given dataset is an interesting research direction, which is  partially explored also in this work.} This can be explained by observing that all systems/channels contribute to the final voting with the same weight. Therefore, the insertion of the worst hypotheses in the ranked list contributes 
to worsen 
rather 
than to improve  global performance. 

In both tasks, segment-level combination  gives significantly better results than system-level combination. In addition, 
the WER increases by augmenting the number of hypotheses 
considered.
Actually, being able to correctly select the 
best  hypothesis for each segment  (SegO - \textit{L1}) we would obtain the highest performance on both tasks, respectively 8.9\% on IWSLT and 19.5\% on CHiME-3. 

The last  two rows of Table~\ref{motivation} give the performance of ROVER when the transcriptions are combined in inverse order, from the worst to the best one, at both system  (InSysO) and segment level (InSegO). The poor results achieved, especially at lower combination levels, demonstrate that correctly ranking the input hypotheses  is essential to improve ROVER combination results. To address this problem, we investigate two different strategies: ranking by regression (RR) and machine-learned ranking (MLR).

\subsection{Ranking by regression (RR)} 
\label{secRR}
This strategy exploits a supervised regression model to predict the WER of each transcription at segment-level. Then, it ranks them from the best (lowest predicted WER) to the worst (highest predicted WER). 
Referring to the structure of Figure~\ref{arcpic}, for training the QE regressor on a training set with $K$ segments, the training set can be defined as:

\noindent
$I= \{ (S_{i}^{k},H_{i,j}^{k},TW_{i,j}^{k}); 1\leq k\leq K,1\leq i\leq M,1\leq j\leq N\}$.

\noindent
where $TW_{i,j}^{k}$ stands for the true WER and it is computed by aligning 
each input hypothesis $H_{i,j}^{k}$ to the manual reference.
The regressor is trained and tuned to predict the true WER following a \textit{k}-fold cross validation scheme. In the experiments reported in the next sections we will refer to this method as RR1.

For the IWSLT task, where there is only one microphone but several ASR systems, we use a second strategy, which we will call  RR2. This method is similar to RR1 but, instead of using all the transcriptions of each segment to train the regressors, it uses only one hypothesis, which is randomly selected. 
The reason is that in IWSLT the speech signal of a segment is shared among different transcriptions, which makes the signal-based features identical, hence uninformative or even misleading for training the regressors. 
In this case,
the training set can be defined as:

\noindent
$I= \{ (S_{i}^{k},H_{i,j}^{k},TW_{i,j}^{k}); 1\leq k\leq K,1\leq i\leq M,j \in rand(1..N)\}$.

\noindent
where $rand(1..N)$ outputs a random number between $1$ and $N$.
Note that we will not use this second RR strategy for the CHiME-3 task, in which the multiple microphones record different signals for each utterance.
In this case, all the signal-based features are 
different and informative for all the resulting hypotheses.

\subsection{Machine-learned ranking (MLR)} \label{secMLR}
This strategy exploits machine-learned ranking algorithms \cite{cao2007learning, mcfee2010metric, clemenccon2013ranking}, which are widely used in information retrieval and question answering tasks. Unlike ranking by regression, in which we first predict the WER and then we use it to estimate the rank of each transcription, here we directly predict the ranking. In this approach the labels of the training instances are  the true ranks:

\noindent
$I= \{ (S_{i}^{k},H_{i,j}^{k},TR_{i,j}^{k}); 1\leq k\leq K,1\leq i\leq M,1\leq j\leq N\}$.

\noindent
where $TR_{i,j}^{k}$ is the corresponding ``true rank" derived from the true WER value ($TW_{i,j}^{k}$). For instance, given two hypotheses ($H_{x,y}^{k}$ and $H_{z,w}^{k}$) and their true WERs ($TW_{x,y}^{k}$ and $TW_{z,w}^{k}$), we consider $TR_{x,y}^{k} \preceq TR_{z,w}^{k}$, if $TW_{x,y}^{k} \leq TW_{z,w}^{k}$. Similarly to the RR approach, the ranking models are trained on feature vectors extracted from 
the signal and the corresponding transcription hypotheses.

{MLR, differently from the regression-based strategy, performs a pairwise comparison between the candidates \cite{cao2007learning}. 
For any pair of segment transcriptions generated by two or more systems , it processes the corresponding feature vectors and decides to place one transcription ahead of the other.
The objective function is designed to minimize the average number of inversions in ranking.}

The learning capability of MLR is affected by the level of separability of the instances to be ranked. Sets of hypotheses characterized by large quality differences will favour this learning algorithm, while higher similarity in the hypotheses (and, consequently, a larger presence of ties) will reduce the possibility  to return correct ranks. Section~\ref{diss} provides an initial investigation on the sensitivity of MLR to the 
number 
of ties in the training data.

\subsection{Features} \label{feat}
To train the
QE
models, from  signals and the corresponding  automatic transcriptions we extract four sets of features: signal, textual, hybrid and word-based. 
{\textbf{Signal} features look at each signal as a whole, trying to capture the difficulty to transcribe it. To this end, they are extracted by analyzing the audio waveform with a window of 20ms length at a frame rate of 10ms.  
\textbf{Textual} features try to model the plausibility (\textit{i.e.} the fluency) of a transcription hypothesis by looking at its content and language model probabilities.
\textbf{Hybrid} features 
are intended to characterize the difficulty of transcribing the signal in a more fine-grained way. 
Also this group is indeed based on signal information, but it also considers as constraint the word time boundaries provided by the decoder at transcription time.
These three groups of 
basic
features,
correspond to those proposed in \cite{negri2014quality}.}
\textbf{Word-based} features are new in ASR QE and they are inspired by previous approaches to ASR word error detection \cite{Chieu:2002:NER:1072228.1072253,pellegrini2010improving,goldwater2010words,tam2014asr}.
They aim to capture word pronunciation and recognition difficulty, which can be respectively obtained by counting the number of homophones/heteronyms (similar pronunciations) and by computing  word-level language model (LM) probabilities.
For the former we use a lexicon and for the latter we use \textit{n}-gram  and recurrent neural network language models \cite{mikolov2010recurrent}.
These language models are trained on the official data provided by IWSLT and CHiME-3 organizers.
Having segments of different length, word-based features are computed by averaging the values of their individual words. The complete list of features is reported in Table \ref{feattab}.

\begin{table*} [h]
\begin{center}
\begin{small}
\begin{tabular}{|p{1.5cm}|p{14cm}|}
\hline

\textbf{Signal} (17) &  mean values of 12 Mel Frequency Cepstral Coefficients 
removing the $0^{th}$ order coefficient is discarded (12), log energy computed on the whole segments (1),  the mean/min/max  values of raw energy (3), total segment duration (1). \\ \hline

\textbf{Textual} (10) & number of words (1), LM log probability (1), LM log probability of part of speech (POS) (1), log perplexity (1), LM log perplexity of POS (1),  percentage (\%)  of numbers (1), \% of tokens  which  do not contain only ``[a-z]" (1), \% of content  words (1), \% of nouns (1), \% of verbs (1). \\ \hline

\textbf{Hybrid} (26) & signal-to-noise ratio (SNR) (1), mean/min/max  noise energy (3), mean/min/max word energy (3),  (max word - min noise) energy (1),  number of silences (\#sil) (1), \#sil per second (1), number of words (\#wrd) per second (1), $\frac{\#sil}{\#wrd}$ (1), total duration of words ($D_{wrd}$) (1), total duration of silences ($D_{sil}$) (1), mean duration of words (1), mean duration of silences (1), $\frac{D_{sil}}{D_{wrd}}$ (1), $D_{wrd}-D_{sil}$ (1), standard deviation (std)  of word duration (1), std of silence duration (1), mean/std/min/max of pitch\footnote{Pitch features have been computed with the Praat software tool \cite{Praat2005}.} (4), number of hesitations (1), frequency of hesitations (1). \\ \hline

\textbf{Word} (22) & POS-tag/score of the previous/current/next words (6), RNNLM   probabilities given by models trained on in-domain/out-of-domain  data (2), in-domain/out-of-domain 4-gram LM probability (2), number of   phoneme classes including fricatives, liquids, nasals, stops and vowels (5), number of homophones (1), number of lexical neighbors (heteronyms) (1) binary   features answering the three questions: ``is the current word a stop word?"/"is the current word before/after repetition?"/"is the current word before/after silence?" (5). \\ \hline

\end{tabular}
\end{small}
\end{center}
\caption{\label{feattab} 
Complete list of the 75 features used to train our ASR QE models.}
\end{table*}

\section{Experimental setting}
\label{experiment}
In this section we describe the data used for
our experiments, 
as well as
the metrics and the  baselines adopted  to evaluate 
our approach.

\subsection{IWSLT}
\label{subsec:tedd}
For this task we use the submissions to the IWSLT2012 and IWSLT2013 ASR evaluation campaigns
respectively
as  
training and test sets.
Both sets are collected from English TED talks dealing with different topics. 
Six groups participated in IWSLT2012, 
in which the best performance (12.4\% WER) was obtained by NICT\footnote{National Institute of Information and Communications Technology, Kyoto, Japan} group. Two more groups participated in IWSLT2013, which was won again by NICT (13.5\%). Tables~\ref{iwsltstat} and \ref{iwsltscores} 
provide 
some basic statistics about this data and participants' WER results.
Complete details about each ASR system can be found in \cite{federico2012overview} and \cite{cettolo2013report}.

By selecting the data from two different years, we guarantee that
there is no speaker nor topic overlap between training and test sets. 
In addition, although most of the submissions of 2013 come from the same laboratories of 2012 (except for two teams  that did not participate in
2012) the ASR systems  are quite different due to the changes and the improvements made
by participants 
during the course of one year of research. Finally, in this audio dataset, since the recordings were carried out by head-mounted microphones,  there is neither reverberation nor background noise apart from the applause breaks,
especially in the final part of the talks. However, it happens frequently that the speakers, sometimes non-native ones, introduce spontaneous speech phenomena such as hesitations, repetitions and false starts during their talks.
\begin{table}
\parbox{.45\linewidth}{
\centering
\small
\begin{tabular}{|l|c|c|}
\hline
\multirow{2}{*}{Attributes} & IWSLT2012 & IWSLT2013  \\ 
                                & (training)   & (test)    \\\hline\hline

duration (hr) & 1h45m & 4h50m  \\ 
\# sent & 1,124 & 2,246  \\
\# token & 19.2k & 41.6k  \\
dict. size & 2.8k & 5.6k \\
\# speakers & 11 & 28  \\ \hline
\end{tabular}
\caption{\label{iwsltstat} Statistics of IWSLT task.} 
}
\hfill
\parbox{.45\linewidth}{
\centering
\small
\begin{tabular}{|l|c|c|}
\hline
  System & IWSLT2012 & IWSLT2013 \\ \hline \hline
  FBK & 16.8 & 23.2 \\ 
  KIT & 12.7 & 14.4 \\
  MITLL & 13.3 & 15.9 \\
  NAIST & -- & 16.2 \\
  NICT & 12.4 & 13.5 \\
  PRKE & -- & 27.2 \\
  RWTH & 13.6 & 16.0 \\
  UEDIN & 14.4 & 22.1 \\ \hline
  Avg. & 13.86 & 18.56 \\\hline
\end{tabular}
\caption{\label{iwsltscores}IWSLT WER[\%] ($\downarrow$) results.}}
\end{table}
In short, the experiments on this task are conducted as follows:
\begin{enumerate}
\item Segmentation of IWSLT2012 
(training) and IWSLT2013 (test) data;
\item Feature extraction from the $<signal,transcription>$ training pairs;
\item Training of RR/MLR models; 
\item Feature extraction from the $<signal,transcription>$ test pairs;
\item Estimation of the  WERs/ranks of the transcribed test segments; 
\item Execution of ROVER 
and concatenation of the resulting outputs; 
\item Computation of the  WER of the concatenated outputs.
\end{enumerate}

While for the  
IWSLT2012 
audio data
manual utterance segmentation is available, for  
IWSLT2013
the segmentation has to be carried out automatically by each individual ASR system before decoding the audio tracks. Since each system produces a different number of segments, it is necessary to align each automatic segmentation with a reference one in order to share the same utterance time boundaries among the different hypotheses. Although in principle  the segmentation given by one randomly chosen  ASR system could be taken as reference, we decided to align each segmentation  with  the reference segmentation provided within the framework of the IWSLT2013 campaign.

 \subsection{CHiME-3}
\label{chime_exp} 
For the multiple microphones task, we use the publicly available data collected for the $3^{rd}$ CHiME challenge.
In this 
scenario, six 
microphones placed on a tablet PC were used to record sentences of the Wall Street Journal corpus, uttered by different speakers in four 
noisy environments (bus, cafe, pedestrian area, and street junction). 
Two types of noisy data are provided for this task: {\em real} data, \textit{i.e.}, speech recorded in four real noisy environments uttered by actual talkers; {\em simulated} data, \textit{i.e}, noisy utterances generated by artificially mixing clean speech data with noisy backgrounds.
In our experiments we use the \texttt{real} subsets.
As 
training data  
we use {\em dt05\_real} (DT05 henceforth), which is formed by 1,640 sentences uttered by four
speakers. As test data we use  {\em et05\_real} (ET05), which is formed by 1,320 utterances uttered by four other speakers. 
In contrast to IWSLT, since each utterance recording shares the same time segmentation across all microphones,  here no automatic segmentation is needed.

The organizers of the challenge also provided two baseline ASR systems employing the Kaldi toolkit \cite{Povey_ASRU2011}.
One uses the traditional Gaussian mixture model ($*-gmm$) and the other uses a state-of-the-art deep neural network ($*-dnn$). The former is trained with the Kaldi recipe prepared for the previous CHiME challenges \cite{chime1_paper,chime2_paper} and the latter is trained with Karel's setup \cite{karel2011} included in
the toolkit.

Tables~\ref{chimestat} and \ref{chimescores} respectively show  some statistics about 
the CHiME-3 data and the WER obtained by the baseline ASR systems over each  of the 5 different channels (microphones). As mentioned in  \S \ref{qe-rover}, one of the microphones (index 2) is not used in the evaluation as it is located on the back of tablet PC, mainly to capture the  background noise.

\begin{table}
\parbox{.45\linewidth}{
\centering
\small
\begin{tabular}{|l|c|c|}
\hline
\multirow{2}{*}{Attributes} & DT05 & ET05   \\ 
                                & (training)   & (test)    \\\hline\hline

duration (hr) & 2h74m & 2h33m  \\
\# sentences & 1,640 & 1,320  \\
\# words & 27.1k & 21.4k  \\
dict. size & 1.6k & 1.3k  \\
\# speakers & 4 & 4 \\ \hline
\end{tabular}
\caption{\label{chimestat} Statistics of CHiME-3 task.} 
}
\hfill
\parbox{.45\linewidth}{
\centering
\small
\begin{tabular}{|l|c|c|}
\hline
  Channels & DT05 & ET05 \\ \hline \hline
  1-dnn & 20.5 & 32.9 \\
  1-gmm & 23.4 & 37.3 \\
  3-dnn & 20.0 & 38.8 \\
  3-gmm & 24.2 & 40.5 \\
  4-dnn & 18.8 & 38.0 \\ 
  4-gmm & 21.5 & 36.5 \\
  5-dnn & 16.7 & 32.6 \\ 
  5-gmm & 18.7 & 33.2 \\
  6-dnn & 16.5 & 34.4 \\
  6-gmm & 19.2 & 34.8 \\ \hline
  Avg. & 19.95 & 35.9 \\ \hline
\end{tabular}
\caption{\label{chimescores}WER[\%] ($\downarrow$) of CHiME-3 channels.}}
\end{table}

Experiments are conducted similarly to the IWSLT task, with the difference that in this case each hypothesis
is generated by the two baseline systems ({\it gmm} and {\it dnn}) processing each individual microphone audio track ($1,3,4,5,6$).

In distant speech recognition, it is common to combine the signals recorded by a microphone array in order to generate an enhanced  signal. As mentioned in  \S \ref{related}, MVDR and DS are two popular enhancement approaches. 
MVDR processing is included in the baseline system provided by CHiME-3 organizers \cite{CHiME3_paper}. This adaptive algorithm minimizes the variance of the resulting signal and it is implemented according to \cite{mestre2003signal}.
Similarly, DS~\cite{brandstein1997} applies a fixed weight for each microphone, as described in \cite{jalalvand2015boosted}.
For both approaches speaker position is derived from a frame-level computation of the Steered Response Power Phase Transform (SRP-PHAT) that estimates the time difference of arrival between microphone pairs  \cite{knapp1976,brandstein2001}, using a 3-dimensional grid. 
The peaks of SRP-PHAT are then tracked over time using the Viterbi algorithm (see also \cite{loesch2010adaptive,blandin2012} for implementation details and a comparison of source localization techniques.) 
We apply both methods to combine the signals from the 5 microphones and then we transcribe
the resulting signals by using the two aforementioned baseline ASR systems. This produces four additional channels whose WERs are reported in Table~\ref{chimeenh}.

\begin{table} [h] 
\begin{center}
\small
\begin{tabular}{|l|c|c|}
\hline
\multicolumn{3}{|c|}{WER of Enhanced Channels} \\
Enhancment-ASR & DT05 & ET05 \\ \hline
mvdr-gmm & 20.3 & 37.1 \\
mvdr-dnn & 17.6 & 33.1 \\ 
ds-gmm & 12.2 & 23.1 \\ 
ds-dnn & 10.4 & 20.5 \\ \hline
\end{tabular}
\end{center}
\caption{\label{chimeenh}  WER[\%] ($\downarrow$) scores of  the enhanced channels.}
\end{table}

As it can be seen, DS processing significantly outperforms MVDR on both 
training and test sets.
Nevertheless, we keep also the 
MVDR hypotheses
for combination, since we believe that 
they provide complementary solutions whose diversity might improve final results.

Comparing the WER scores reported in Tables~\ref{iwsltscores} and \ref{chimescores}, it is worth noting that the quality of the CHiME-3 transcriptions is globally lower
and, on average, training and test data are 
more distant. As we will discuss in  \S \ref{results}, this can explain why:  \textit{i)} the performance achieved by  standard random ROVER is lower on CHiME-3 data than on IWSLT, \textit{ii)} adding high quality transcriptions to the combination   (\textit{e.g.} those obtained from signal enhancement) results in much larger WER reductions on  CHiME-3, and \textit{iii)} ranking methods seem to suffer from the presence of a large number of transcriptions with very similar WER scores.

\subsection{Terms of comparison}
In both sets of experiments we compare our approach with the standard random ROVER 
and the two oracles described in  \S \ref{qe-rover}. In random ROVER the 
entire
transcription candidates obtained from different systems/channels are taken in random order.
In the system-level oracle (SysO) we assume to know 
their true 
overall
rank, 
which is kept unchanged for all the segments.
Moving to a finer granularity, 
in the segment-level oracle (SegO) we assume to know the exact rank of the ASR hypotheses at the segment level and  ROVER is applied segment by segment using the known 
ranks. The result obtained in this way 
is
our strongest term of comparison.

We consider the last two methods as ``oracles'' as they rely on external information about the actual system ranks, which is not accessible to the QE-informed approach that we aim to evaluate. 
Although our first goal is to significantly outperform  random ROVER, reducing the performance gap with respect to the two oracles would hence represent an even more convincing measure of success.

\subsection{Evaluation metrics}
All our results are computed by measuring word error rate. WER  is computed by dividing the total number of errors (substitutions, insertions and deletions) by the total number of reference words.

\begin{equation*}
WER = \frac{\#substitutions + \#insertions + \#deletions}{\#reference\_words}
\end{equation*}

The regressor parameters are tuned to minimize mean absolute error (MAE) between  the predicted WER ($predWER$) and 
true 
WER ($trueWER$). For $K$ segments, recorded by $M$ microphones and transcribed by $N$ ASR systems, MAE is computed as:

\begin{equation*}
MAE = \frac{1}{K\times M\times N}\times \sum_{k=1}^{K} \sum_{i=1}^{M} \sum_{j=1}^{N}  \left | predWER_{i,j}^{k}-trueWER_{i,j}^{k}  \right | 
\end{equation*}

To tune the parameters of the ranking models we maximize
the  mean average precision ($MAP@L$) of the predicted ranks
at each level of combination ($L=M\times N$).
This 
quantity is defined as follows:

\begin{equation*}
MAP@L=\frac{1}{K}\sum_{k=1}^{K}AP^{k}@L
\end{equation*}
where $K$ is the total number of segments and $AP^{k}@L$ is the average rank precision of the $k^{th}$ segment. $AP^{k}@L$ is computed as follows:

\begin{equation*}
AP^{k}@L=\frac{\sum_{l=1}^{L}\frac{P(l)}{l}}{L}
\end{equation*}
where $P(l)$ is the number of correct ranks at level $l$.

To perform significance tests,
we 
run 
the matched-pairs test \cite{gillick1989some}  at a significance level of 95\% (\textit{i.e.}  $p=0.05$). This method randomly selects subsets of segments processed with two given combination approaches and calculates the corresponding WERs in each subset. If the percentage of  agreements between the two  approaches is above 95\%, then they are not considered as significantly different. In order to simplify the comparison between different combination methods in the result tables, we use the three following symbols:

\begin{enumerate}
\item ``$\dagger$'' indicates that the WER score obtained
after running ROVER with a given ranking method
is not significantly different from random ROVER;
\item ``$\bullet$'' indicates that the WER score obtained 
after running ROVER with a given ranking method
is not significantly different from the system-level oracle;
\item ``$\star$'' indicates that the WER score obtained 
after running ROVER with a given ranking method
is not significantly different from the  segment-level  oracle.
\end{enumerate}

Based on 
the above definitions
``$\dagger$'' indicates a negative result, as it is close to the baseline random ROVER, while ``$\bullet$'' and ``$\star$'' indicate 
a positive result, close to the oracles.

\subsection{Ranking setup}

In order to implement the first ranking strategy, \textit{i.e.} ranking by regression (RR), we
rely on the findings of \cite{negri2014quality}, which indicates that  extremely randomized trees (XRT) \cite{geurts2006extremely} is a suitable regression algorithm for this task.  XRT is a combination of several small trees, each trained on the whole data but with a random subset of features. Afterwards, similarly to other ensemble classification methods, an ensemble approach is used for the fusion of the output of all these small trees. For our experiments we use the implementation provided in the Scikit-learn package \cite{pedregosa2011scikit}. The parameters like the number of extra trees, the number of features per tree and the number of instances in the leaves are tuned to minimize MAE, using \textit{k}-fold cross-validation on the training set.

For the second strategy, \textit{i.e.} machine learned ranking (MLR), we adopt the solution proposed in  \cite{sjalalvand-EtAl:2015:ACL}, which identifies in  the random forest algorithm \cite{breiman2001random,clemenccon2013ranking} the best ranking model for this kind of tasks. In particular, we use the implementation provided as an option in the RankLib library.\footnote {\url{http://sourceforge.net/p/lemur/wiki/RankLib}}
The main parameters, such as the number of bags, the number of trees per bag and the number of leaves per tree are tuned to maximize MAP, using \textit{k}-fold cross-validation.

All models are trained with three different feature settings: \textit{i)} signal, textual and hybrid features, indicated in the tables below with ``\textit{+B}''; \textit{ii)} word-based features, indicated with ``\textit{+W}''; and \textit{iii)} the combination of both, indicated with ``\textit{+BW}''.

\section{Results}
\label{results}
For both IWSLT and CHiME-3 tasks, we run our QE-informed ROVER approach at different levels of combination, that is with different numbers of ASR components contributing to the 
fusion process.
This number is always higher than 3, since a lower value would be meaningless when majority voting is applied. Therefore, the combination level for IWSLT ranges in the interval $[L3,L8]$, while for CHiME-3 it ranges in $[L3,L14]$.

It's worth remarking that the results reported in this work are not comparable with the official ones of 
CHiME-3, since here we assume that 
neither the ASR systems nor their confidence scores are accessible. Instead, the official submissions (like our own ones described in \cite{jalalvand2015boosted}) take advantage of a combination of acoustic model adaptation, confidence-based data selection and multi-pass recognition as well as signal processing and language model rescoring techniques.
Here, we 
do not consider
these techniques and we apply our QE approach to the first pass decoding output of the baseline systems. 
This is to confirm that our  
method 
is 
general enough
to be applied 
in a variety of 
``black-box''
conditions.

\subsection{IWSLT}
\label{subsec:iwsltResults}
\paragraph{Preliminary analysis} To test the performance of our method in a 
controlled setting, we first run QE-informed ROVER on the 
training set (IWSLT2012) using 4-fold cross validation. 
Data partitioning is done with regard to the speakers in order to guarantee that there is no speaker overlap between training and 
test folds.
Moreover, the partitioning is done in such a way that each instance occurs only once in each 
test part.
Thus,
by collecting the results from all  
folds, we obtain the performance on whole training data reported in Table \ref{weriw-dev}.

\begin{table*} [t]
\begin{center}
\small
\begin{tabular}{|l|llll|c|}
\hline
\textbf{Ranking}  &L3 & L4 & L5 & L6 & Avg. Impr.\\ \hline\hline
Random & 13.4 & 11.8 & 12.3 & 11.8 & 0.0\\ 
SysO & 11.4 & 9.3 & 9.6 & 9.5 & -2.4 \\
SegO & 8.0  & 7.9 & 8.2 & 9.1 & -4.0\\ \hline \hline
RR1+B & 11.5 & 10.2 & 9.9 & 9.8 & -2.0 \\
RR1+W & 13.2 & 10.6 & 10.1 & 9.8 & -1.4 \\
RR1+BW & 12.1 & 10.3 & 10.0 & 9.8 & -1.8 \\ \hline
RR2+B & 11.3 & 10.3 & 9.9 & 9.8 & -2.0\\
RR2+W & 12.1 & 10.3 & 10.0 & 9.8 & -1.8 \\
RR2+BW & 11.2 & 9.9 & 9.8 & \textbf{9.6} & -2.2 \\ \hline
MLR+B & 10.7 & 9.8 & 9.7 & \textbf{9.6} & -2.4 \\
MLR+W & 10.7 & \textbf{9.7} & 9.7 & \textbf{9.6} & -2.4 \\
MLR+BW & \textbf{10.6} & 9.8 & \textbf{9.6} & \textbf{9.6} & -2.4 \\ \hline
\end{tabular}
\end{center}
\caption{\label{weriw-dev} WER[\%] ($\downarrow$)  of different ranking methods on IWSLT2012 using 4-fold CV.}
\end{table*}

The first three rows of the table show the results of ROVER when the input hypotheses are randomly ordered at system level, and those achieved 
by the two oracles previously described (SysO and SegO). 
Even if random ROVER achieves better results than the individual IWSLT2012 participants (see Table~\ref{iwsltscores}), it is quite far from the oracles.
SysO, for which the overall ranking of the ASR systems is known and  is kept unchanged for all the segments, is significantly  better, with an average WER reduction of 2.4\% over random ROVER. As expected, the best performance is achieved by the segment-level oracle with an average WER reduction 
of 4.0\%. These differences confirm that running ROVER without an optimal order of the hypotheses may limit the  gain achievable by system combination.
Apart from that, 
both oracles reach the lowest WER score at combination level $L4$. This indicates that the majority of aligned words in the confusion sets of the word network built by ROVER falls in the top four 
hypotheses.
Inserting more alternative words from less accurate hypotheses at $L5$ and $L6$ has a negative impact on majority voting.

Ranking by regression, using all the training transcriptions and
the basic  
features (RR1+B) already outperforms random ROVER significantly but 
remains far from the oracles.
Note that basic features include signal, textual and hybrid features from Table \ref{feattab}.
Slightly worse performance is obtained when using the word-based features (RR1+W) and the union of the features (RR1+BW),  suggesting that the word features are not particularly useful in this setting. 

Although 
the various ASR systems provide different hypotheses for each decoded segment, the corresponding signal feature vectors are the same and, therefore, the trained regressor could be misled by the co-occurrence of similar feature values with different training labels. To overcome this problem, as explained in  \S \ref{secRR} we trained regression models (called RR2 in Table~\ref{weriw-dev}) by using only one
random 
transcription for each segment.
This approach, using both basic and word-based features (RR2+BW), yields the best performance among the RR methods (on average, the WER is reduced by 2.2\%). 
Interestingly, 
both RR1 and RR2 reach the best performance at combination level $L6$. This behavior, differently from the oracle conditions, indicates that the majority of the words in each confusion set of the ROVER word network is not concentrated on the top ``automatically" ranked positions but is more uniformly distributed over all combination levels.
Machine-learned ranking with the basic features (MLR+B), word-based features (MLR+W)  and their combination (MLR+BW) always exhibits 
the largest WER reductions over random ROVER (-2.4\% on average).
Its results are not only consistently better than RR, but also close to system-level oracle (SysO).

In summary, this preliminary analysis
suggests that: \textit{i)} regardless of the QE method and the features used, QE-informed ROVER outperforms random ROVER with a  large margin, \textit{ii)} among all the QE strategies, MLR+BW shows the best performance at most of the levels, and \textit{iii)} though still far from the strongest term of comparison (SegO), 
its results  are competitive with those of the system-level oracle.
As shown in the last column of Table~\ref{weriw-dev}, the average improvement of the two methods are almost the same (-1.3\% vs. -1.1\%).

\paragraph{Test}

Using the QE models trained on IWSLT2012, we carry out the same set of experiments on the test data. The corresponding results are reported in Table~\ref{weriw-tst}. Since in IWSLT2013 there are 8 teams who submitted their transcriptions, in this case the combination level varies in [$L3,L8$] interval.

\begin{table*} [h]
\begin{center}
\small
\begin{tabular}{|l|llllll|c|}
\hline
\textbf{Ranking}  &L3 & L4 & L5 & L6 & L7 & L8 & Avg. Impr\\ \hline\hline
Random &14.6 & 13.7 & 13.2 & 12.8 & 12.7 & 12.4 & 0.0 \\ 
SysO   &12.2 & 11.7 & 11.8 & 11.9 & 12.1 & 12.1 & -1.3 \\ 
SegO   &10.5 & 11.0 & 11.4 & 11.6 & 11.7 & 11.7 & -1.9 \\ \hline \hline
RR1+B  &13.9 & 13.1 & 12.6 & 12.4 & 12.4 & 12.3 $\dagger$ $\bullet$ & -0.4 \\
RR1+W  &14.0 & 13.0 & 12.5 & 12.2 & 12.3 $\bullet$ & 12.3 $\dagger$ $\bullet$ & -0.5 \\
RR1+BW &14.0 & 13.0 & 12.5 & 12.2 & 12.3 $\bullet$ & 12.3 $\dagger$ $\bullet$ & -0.5 \\ \hline
RR2+B  &13.8 & 13.0 & 12.6 & 12.4 & 12.3 $\bullet$ & 12.3 $\dagger$ $\bullet$ & -0.5 \\
RR2+W  &14.2 & 13.1 & 12.7 & 12.4 & 12.5 $\dagger$  & 12.4 $\dagger$ $\bullet$& -0.3 \\
RR2+BW &13.7 & 12.8 & 12.4 & 12.2 & \textbf{12.2} $\bullet$ & \textbf{12.2} $\dagger$ $\bullet$ & -0.6\\ \hline
MLR+B  &12.9 & 12.4 & 12.3 & 12.1 $\bullet$ & 12.3 & \textbf{12.2} $\dagger$ $\bullet$ & -0.9\\ 
MLR+W  &\textbf{12.4} $\bullet$ & \textbf{12.1} & \textbf{12.0} & 12.0 $\bullet$ & \textbf{12.2} $\bullet$ & \textbf{12.2} $\dagger$ $\bullet$ & -1.1 \\
MLR+BW & \textbf{12.4} $\bullet$ & \textbf{12.1} & \textbf{12.0} $\bullet$ & \textbf{11.9} $\bullet$ $\star$ & \textbf{12.2} $\bullet$  & \textbf{12.2} $\dagger$ $\bullet$ & -1.1 \\ \hline
\end{tabular}
\end{center}

\caption{\label{weriw-tst} WER[\%] ($\downarrow$)  of different ranking methods on 
IWSLT2013.
 ``$\dagger$'' =  the result is not statistically different from random ROVER;  ``$\bullet$'' = the result is not statistically different from SysO; ``$\star$'' the result is not statistically different from SegO.}

\end{table*}

Although the overall trend is similar to the one observed on 
the training 
data, the improvement margins are smaller. The reason for this unsurprising difference is the mismatch between training and test data. In fact, in our preliminary experiments, the same set of ASR systems (6 systems) was used both in training and test. Here, 
instead,
the ASR systems are different in number and nature (6 systems from IWSLT2012 for training, 8 systems from IWSLT2013 for test). Nevertheless, in many columns of Table~\ref{weriw-tst} the WER reduction is large and significant. The best result (MLR+BW at $L6$), in particular, is not only better than random ROVER, but also not statistically different from the strongest   oracle (SegO). Also on the test set, MLR seems to work better than RR in ranking the input hypotheses. One possible reason is the higher reliability of pairwise comparisons compared to 
considering numeric scores that reflect independent quality predictions.

So far, we showed that: \textit{i)}  we can improve random ROVER using ASR QE for ranking the inputs, and \textit{ii)} with an appropriate QE approach and an efficient set of features, we can  approach the strong system-level and segment-level oracles. The scores reported in Table~\ref{weriw-tst}, however, only provide a global picture that might hide interesting details such as larger gains in specific conditions 
that are particularly  favorable for QE-based ranking.

\begin{figure}
\centering
  \centering
  \includegraphics[trim=1cm 0.0cm 0.5cm 0cm, clip=true,width=0.55\textwidth]{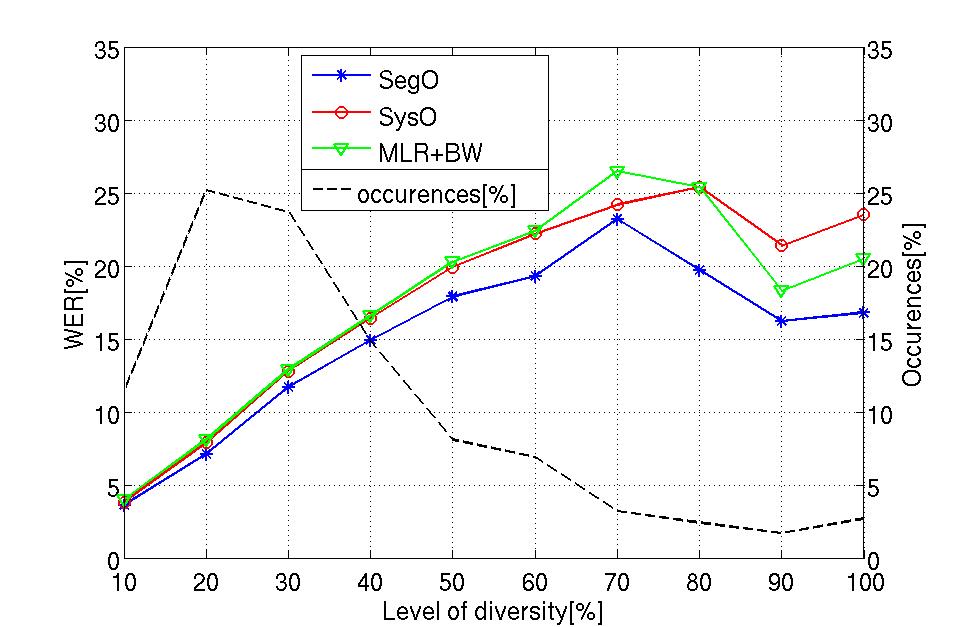}
  \caption{Performance as function of hypothesis diversity  ($div=MAX_{WER}[\%]-MIN_{WER}[\%]$) in IWSLT.
  }
  \label{chanal11}
\end{figure}

To better investigate this aspect, in Figure~\ref{chanal11}, we analyse the performance on different groups of segments characterized by different levels of diversity. By diversity, we refer to the level of disagreement between ASR systems. It is computed as the difference between the maximum and minimum 
WERs among the transcriptions of a given segment. With this definition, we divide the segments into 10 groups (X-axis in Figure~\ref{chanal11}). The first group consists of the segments whose transcriptions diversity is lower than 10\% (\textit{i.e.} the WER difference between the best and the worst  transcription is less than 10\%). In the second group, the diversity is between 
[10\%,20\%] and so on. 
The figure reports the performance of ROVER using SegO, SysO and MLR+BW over each diversity 
group, as well as the proportion of segments belonging to each group (black-dotted line).
As it can be seen, most of the segments have low diversity (less than 20\%), meaning that the ASR systems 
return
similar transcriptions for them. It is interesting to note  how the relative differences between the three systems are affected by hypothesis diversity. For low values, in which ranking is intuitively more difficult and  less accurate, MLR+BW's performance is less reliable and close to SysO. 
{For high diversity values (\textit{i.e.} above 70\%), in which the ranker} is able to order the components more precisely, MLR+BW outperforms SysO halving the gap that separates it from SegO in the last two bins. This interesting trend is hidden by the fact that the IWSLT data are characterized by low diversity in the transcriptions (highly-diverse hypotheses can be found for less than 7\% of the segments). This, in turn, results in global WER scores where large gains on few segments are smoothed by small gains on many segments. We hypothesize that, though suitable to evaluate our approach, the IWSLT scenario is not the ideal one to fully appreciate its potential. In the next set of experiments we apply QE-informed ROVER to the CHiME-3 data, in which the diversity among the microphone channels is higher. If our hypothesis is correct, then larger WER gains over SysO and further reductions in the distance from SegO  will be observed.

\subsection{CHiME-3}
\label{chime3sec}
\paragraph{Preliminary analysis} Differently from the IWSLT task, where the transcriptions  come from multiple ASR systems, in CHiME-3 the transcriptions  come from multiple channels and two ASR systems. Also in this task we first train the QE models on the 
training set (DT05) 
using 8-fold cross validation. The partitioning is done to avoid speaker or sentence overlaps between  training and test 
folds. 
As mentioned in  \S \ref{chime_exp}, for each utterance there are five signals recorded by the microphones, plus two enhanced signals. Each signal is transcribed using both  GMM and DNN acoustic models. Hence, 14 hypotheses are computed for each segment and, consequently, the combination levels range in the interval [\textit{L3,L14}].
\begin{table*} [h]
\begin{center}
\begin{small}
\begin{tabular}{|l|p{0.6cm}p{0.6cm}p{0.6cm}p{0.6cm}p{0.6cm}p{0.6cm}p{0.6cm}p{0.6cm}p{0.6cm}p{0.6cm}p{0.6cm}p{0.6cm}|c|}
\hline
\multirow{2}{*}{\textbf{Ranking}}  & \multirow{2}{*}{L3} & \multirow{2}{*}{L4} & \multirow{2}{*}{L5} & \multirow{2}{*}{L6} & \multirow{2}{*}{L7} & \multirow{2}{*}{L8} & \multirow{2}{*}{L9} & \multirow{2}{*}{L10} & \multirow{2}{*}{L11} & \multirow{2}{*}{L12} & \multirow{2}{*}{L13} & \multirow{2}{*}{L14} & Avg. \\
& & & & & & & & & & & & & Impr. \\ \hline \hline
Random & 14.3 & 13.4 & 12.7 & 12.4 & 12.1 & 11.9 & 11.9 & 11.8 & 11.8 & 11.8 & 11.8 & 12.2 & 0.0 \\ 
SysO & 11.0 & 10.9 & 10.3 & 10.4 & 10.6 & 10.6 & 10.9 & 10.7 & 10.9 & 10.9 & 11.2 & 11.5 & -1.5 \\
SegO & 6.8 & 7.1 & 7.5 & 7.8 & 8.3 & 8.6 & 9.2 & 9.4 & 10.0 & 10.4 & 10.9 & 11.2 & -3.4 \\ \hline \hline
RR1+B & 10.6 & 10.3 & 10.0 & 9.9 & 10.0 & 10.1 & 10.3 & 10.4 & 10.8 & 11.0 & 12.0 & 12.0 & -1.7 \\
RR1+W & 12.3 & 11.4 & 10.9 & 10.7 & 10.7 & 10.7 & 10.8 & 11.1 & 11.3 & 11.7 & 12.0 & 12.1 & -1.0 \\
RR1+BW & 10.3 & 10.0 & 9.7 & 9.7 & 9.9 & \textbf{9.9} & \textbf{10.2} & 10.4 & 10.7 & \textbf{10.0} & 11.7 & \textbf{11.9} & -2.0 \\ \hline
MLR+B & 10.0 & 9.7 & 9.6 & 9.6 & 9.8 & 10.0 & \textbf{10.2} & 10.4 & \textbf{10.5} & 10.7 & \textbf{11.3} & \textbf{11.9} & -2.0 \\
MLR+W & 10.7 & 10.5 & 10.3 & 10.3 & 10.3 & 10.5 & 10.7 & 11.0 & 11.3 & 11.5 & 11.6 & 12.0 & -1.4 \\
MLR+BW & \textbf{9.8} & \textbf{9.5} & \textbf{9.5} & \textbf{9.5} & \textbf{9.7} & \textbf{9.9} & \textbf{10.2} & \textbf{10.3} & \textbf{10.5} & 10.8 & 11.5 & \textbf{11.9} & -2.1 \\ \hline
\end{tabular}
\end{small}
\end{center}
\caption{\label{werch-dev} WER[\%] ($\downarrow$)  of different ranking methods on 
the CHiME-3  training set (DT05)  using 8-fold CV. }

\end{table*}

In this task we do not take into account the RR2 approach. Indeed, since the microphones that record the signal are positioned in different places over the tablet PC, each of them captures a different signal, thus making all the signal-based features different and informative.

The results reported in Table \ref{werch-dev} show that, similar to the IWSLT task,  the best performance is achieved in most of the cases by MLR+BW (with an average WER reduction of 2.1\%  over random ROVER). Unlike IWSLT, most of our results outperform the SysO oracle,
which always uses the enhanced channels as the first input. The best WERs are usually obtained at the lower levels (\textit{i.e.} 9.5\% at $L4$, $L5$ and $L6$). 
This can be explained by
the large gap between the best channels ($ds-dnn$ and $ds-gmm$, respectively  obtaining 10.4\% and 12.2\% WER) and the other channels (obtaining from  16.5\% to 24.2\% WER). By adding more hypotheses,  worse transcriptions are considered in the combination and, consequently, majority voting makes more mistakes. At $L13$ and $L14$, indeed, our results remain slightly worse than SysO.

An interesting achievement with this approach is the significant improvement of hypothesis-level combination with respect to the signal-level combination.
The top result obtained by MLR+BW (9.5\%) is indeed better than the best one reported in Table \ref{chimeenh} for signal-level combination (10.4\% with $ds-dnn$).
In contrast with previous works suggesting that hypothesis-level combination does not yield any significant improvement on  top of signal-level combination \cite{stolcke2011making}, our results show that better final results  can be obtained
through an accurate ordering of the inputs.

\paragraph{Test} Table~\ref{werch}   
shows
the WER results achieved when QE models are trained on the 
DT05 and then used to predict the quality of the 
ET05 transcriptions. The observations of our preliminary analysis seem to be confirmed.  Although trained and tested on different data, QE-informed ROVER  consistently outperforms  random ROVER and the SysO oracle at all levels except for $L14$, where the WER  
variations
achieved by our best system are not statistically significant. 
This supports our hypothesis that, if the diversity among the components is high enough, then the proposed segment-level QE-informed ROVER can lead to better results than the 
system-level oracle.
SegO performance is still significantly better 
at all levels, especially the lower ones where the transcriptions obtained from the best enhanced channels are  hard to  improve with hypotheses coming from the other channels.

It is important to remark that the QE-informed ROVER also significantly improves the performance of the enhanced channels when combining less than eight transcriptions. In fact,  as shown in Table~\ref{chimeenh}, the enhanced $ds-dnn$ channel 
achieves 
20.5\% WER
on 
ET05,
while RR1+BW reduces the error down to 19.1\% at $L5$. As mentioned before, probably due to the high performance difference between the enhanced  and 
raw
channels, we do not observe the same gain at higher levels.

\begin{table*} [h]
\begin{small}
\begin{center}
\begin{tabular}{|l|p{0.6cm}p{0.6cm}p{0.6cm}p{0.6cm}p{0.6cm}p{0.6cm}p{0.6cm}p{0.6cm}p{0.6cm}p{0.6cm}p{0.6cm}p{1.25cm}|c|}
\hline

\multirow{2}{*}{\textbf{Ranking}}  & \multirow{2}{*}{L3} & \multirow{2}{*}{L4} & \multirow{2}{*}{L5} & \multirow{2}{*}{L6} & \multirow{2}{*}{L7} & \multirow{2}{*}{L8} & \multirow{2}{*}{L9} & \multirow{2}{*}{L10} & \multirow{2}{*}{L11} & \multirow{2}{*}{L12} & \multirow{2}{*}{L13} & \multirow{2}{*}{L14} & Avg. \\
& & & & & & & & & & & & & Impr. \\ \hline \hline
Random & 28.7 & 27.4 & 27.2 & 26.8 & 26.5 & 26.4 & 26.2 & 26.2 & 26.1 & 26.1 & 26.2 & 26.3 & 0.0\\ 
SysO   & 22.7 & 22.0 & 21.6 & 21.1 & 21.9 & 22.0 & 24.0 & 24.1 & 24.4 & 24.9 & 25.5 & 25.8 & -3.3\\ 
SegO   & 14.8 & 15.4 & 16.2 & 17.0 & 18.1 & 18.7 & 19.7 & 20.3 & 21.1 & 21.9 & 22.8 & 23.6 & -7.4\\ \hline \hline
RR1+B  & 20.4 & \textbf{19.5} & 19.5 & 20.1 & 20.2 & 20.6 & 21.1 & 21.7 & 22.2 & \textbf{22.9} & 24.2 & 25.8$\dagger\bullet$ & -5.1\\
RR1+W  & 22.2$\bullet$ & 21.1 & 20.3 & 20.3 & 20.5 & 21.0 & 21.3 & 22.0 & 22.9 & 24.0 & 24.9 & 26.0$\dagger\bullet$  & -4.5\\
RR1+BW & 20.0 & \textbf{19.5} & \textbf{19.1} & \textbf{19.5} & \textbf{19.7} & \textbf{20.3} & \textbf{20.7} & \textbf{21.4} & \textbf{22.1} & \textbf{22.9} & \textbf{23.9} & 25.8$\dagger\bullet$ & -5.4\\ \hline
MLR+B  & 20.4 & 19.9 & 19.6 & 20.0 & 20.3 & 20.6 & 21.2 & 21.5 & 22.5 & 23.3 & 24.9 & 25.8$\dagger\bullet$& -5.0\\
MLR+W  & 22.4$\bullet$ & 21.4 & 20.8 & 20.7 & 21.1 & 21.5 & 22.0 & 22.6 & 23.2 & 24.0 & 25.2 & 25.9$\dagger \bullet$& -4.1\\
MLR+BW & \textbf{19.8} & \textbf{19.5} & 19.5 & 19.7 & 20.2 & 20.4 & 20.9 & 21.5 & 22.2 & 23.4 & 24.9 & \textbf{25.7}$\bullet$& -5.2\\\hline
\end{tabular}
\end{center}
\end{small}
\caption{\label{werch} WER[\%] ($\downarrow$)  of different ranking methods on the CHiME-3 test set (ET05).  ``$\dagger$'' =  the result is not statistically different from random ROVER;  ``$\bullet$'' = the result is not statistically different from SysO; ``$\star$'' the result is not statistically different from SegO.}
\end{table*}

Differently from 
IWSLT, where there is an increment of around 2 WER points when moving from the preliminary analysis to the test results, in CHiME-3 we observe a larger degradation ($\sim$10.0\% WER 
on average). 
This is not surprising if we look at the differences in WER between training and test data in the two tasks (Tables \ref{iwsltscores} and \ref{chimescores}), which indicate that 
CHiME-3 data is more heterogeneous.
However,
though working in a more complex scenario,  the QE-informed ROVER is still able to achieve competitive results. 
In terms of learning approach, 
in most of the cases
the best performance on CHiME-3 is obtained  by RR1.
This is in contrast with the IWSLT results, where MLR always performed  better than RR. The main reason for this is related to the fact that in CHiME-3, apart from the enhanced channels, the other channels are quite similar in performance and, overall, they generate transcriptions of low quality with similar or equal WERs (see Table \ref{chimescores}). For MLR, this results in a large number of ties when computing the pairwise comparisons. The  impact on MLR performance with a proper management of ties is addressed in  \S \ref{Ties}. 

\begin{figure}
  \centering
  \includegraphics[trim=1cm 0.0cm 0.5cm 0cm, clip=true,width=0.55\textwidth]{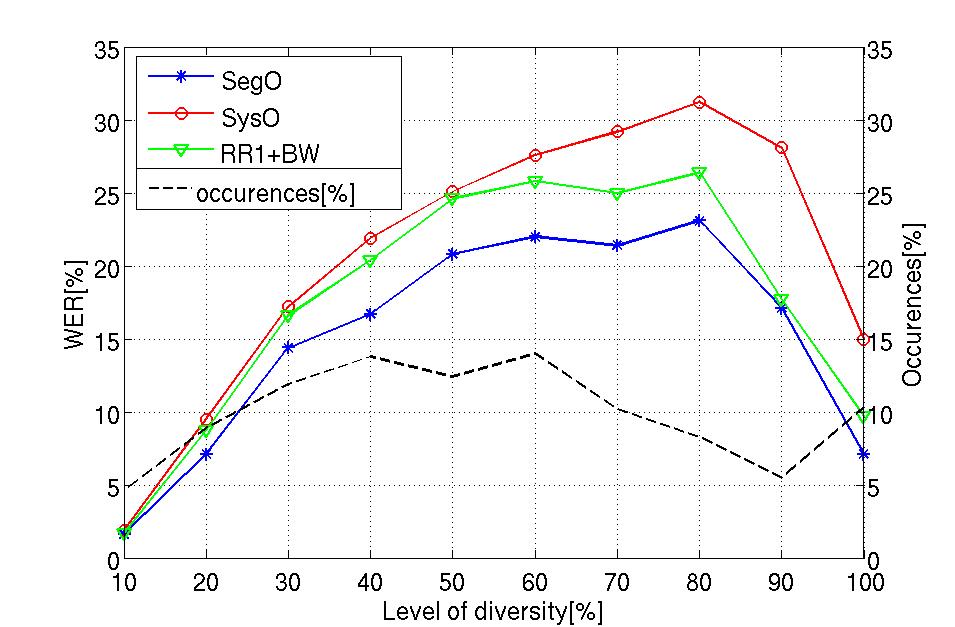}
\caption{Performance as function of hypothesis diversity 
in CHiME-3.
}
\label{chanal12}
\end{figure}

Figure~\ref{chanal12} shows the performance analysis with regard to the diversity groups in the CHiME-3 task. Comparing the curves in Figures \ref{chanal11} and \ref{chanal12}, we observe that the distribution of the CHiME-3 transcriptions with respect to the different levels of diversity (black-dotted lines) is more uniform than in  IWSLT. This is mostly due to the presence of the enhanced channels that perform significantly better than the 
raw ones, hence they enlarge the gap between the best and the worst transcriptions. In terms of performance, this uniform distribution allows the QE-informed ROVER to outperform SysO also for small values of diversity ($\sim$30\% ) 
in 
CHiME-3.
Increasing the level of diversity, our method significantly gains in performance over
 SysO and also approaches SegO at the diversity value of 90\%.

In this section we have shown that: \textit{i)} QE-informed ROVER is able to achieve better performance than  standard ROVER on two very different datasets, \textit{ii)} our approach also outperforms the system-level oracle in the CHiME-3 task, and it obtains competitive results with the enhanced channels, \textit{iii)} the extent of our gains depends on the level of diversity and accuracy of the transcriptions and \textit{iv)} the smallest WER scores are obtained by combining a limited number of transcriptions. One can question how the optimal number of transcriptions can be found in a real unsupervised scenario where the references to compute the WER are not available. We will answer this question in the next section, after analyzing the impact on performance of tied ranks.

\section{Practical issues}
\label{diss}
Despite the positive results discussed so far, our analysis still leaves two important questions open:

\begin{enumerate}
\item Why machine-learned ranking (MLR+BW) shows better performance on IWSLT, whereas in CHiME-3 ranking by regression (RR1+BW) 
generally 
works better?
\item Considering that the best performance is always obtained at a 
task-specific level of combination, how can we predict this level?
\end{enumerate}

In this section we address these two practical issues, respectively by: \textit{i)} analyzing the impact of tied ranks (transcriptions  with  identical  WERs) in  the CHiME-3 training data and \textit{ii)} exploring a classification method to identify the optimum combination level.

\subsection{Tied ranks}
\label{Ties}
In the previous section we observed that the highest results on  IWSLT and CHiME-3 data are 
respectively obtained by MLR and RR. 
This raises the following practical issue: 
{how to find a unique, best performing strategy, suitable for the general case?} 

One 
reason for 
the observed 
difference can be found in the way the QE models are trained. In RR they are trained to separately predict the WER of each transcription, while in MLR they are trained to 
predict relative ranks through pairwise comparisons.
Training sets characterized by a large number of ties 
(transcriptions of the same segment with identical WERs but different quality)
may influence the performance of MLR  more than RR.\footnote{This problem is widely explored in the information retrieval field, where ties are either arbitrarily broken \cite{furnkranz2003pairwise} or managed with \textit{ad-hoc} strategies \cite{Zhou:2008:LRT:1390334.1390382}.}

Table~\ref{ties} reports the average percentage of ties 
for each dataset, showing that IWSLT has less ties 
than CHiME-3
both in 
the
training and test sets.
This seems to contradict the plots of Figure~\ref{chanal12}, which indicate a higher diversity in the CHiME-3 data (compared to IWSLT, instances are more evenly distributed across all diversity levels). Such diversity, however, is 
only due to the presence of the enhanced channels. These are far better than the 
other 
ones, whose 
performance is uniformly 
lower 
and leads to a large number of similar or identical transcriptions. 
Moreover, the CHiME-3  training set includes 8\% more ties than the test set. This suggests that the presence of a large number of ties in the training set of CHiME-3 can be critical and, in turn, it can determine the lower results achieved by MLR.

\begin{table} [h] 
\begin{center}
\small
\begin{tabular}{|l|c|c|}
\hline
 &  IWSLT & CHiME-3 \\ \hline
Train & 38.4\% & 58.9\% \\ \hline
Test & 40.5\% & 50.6\% \\ \hline
\end{tabular}
\end{center}
\caption{\label{ties}  Percentage of ties in each dataset.}
\end{table}

To validate this hypothesis, in the following experiments we break the ties in the training set by looking at the global performance of each 
system/channel
on the same data.\footnote{This information is similar to the prior knowledge that the SysO oracle acquires on the training data and exploits to rank the test hypotheses. Here, however, we fairly operate only on the training set.} 
In particular, if two systems, $A$ and $B$, achieve the same WER on a given training segment, and system $A$ 
has
shown to perform better than system $B$ on the whole training set, then the hypothesis suggested by $A$ will be 
prioritized when breaking the ties.

Table~\ref{wertie}  shows how performance changes if the ties in the training data are broken using such prior knowledge.
In IWSLT, the MLR approach trained on untied ranks ($MLR+BW+Untied$) 
achieves minor improvements. This is due to the fact that:
\textit{i)} the number of ties is not so high to represent a critical issue as we saw in Table \ref{ties},  and 
\textit{ii)} the small WER differences between systems and combination levels  reduce the room for improvement. 
 In CHiME-3, instead,  the model learned from the untied training set ($MLR+BW+Untied$) significantly outperforms $MLR+BW$ at all levels of combination and also outperforms $RR1+BW$ at most of the levels. 
The considerable WER result of 18.6\% obtained by this method at $L4$, which also corresponds to a relative improvement of 9.6\% over the best enhanced channel, represents our best result on CHiME-3. This confirms the validity of our intuition about the negative impact
of tied ranks on MLR performance.
The largest improvement 
is obtained when combining all the transcriptions ($L14$). This is not surprising because this is the condition where more ties  occur.
Interestingly, at this level, the results of QE-informed ROVER improve up to 
the
point that they are no longer statistically different from the strong segment level oracle (see Table~\ref{werch}).

\begin{table*} [h]
\begin{small}
\begin{center}
\begin{tabular}{|l|p{0.6cm}p{0.65cm}p{0.7cm}p{0.8cm}p{0.6cm}p{0.6cm}p{0.55cm}p{0.55cm}p{0.55cm}p{0.55cm}p{0.5cm}p{1.25cm}|c|}
\hline

\multirow{2}{*}{\textbf{IWSLT}}  & \multirow{2}{*}{L3} & \multirow{2}{*}{L4} & \multirow{2}{*}{L5} & \multirow{2}{*}{L6} & \multirow{2}{*}{L7} & \multirow{2}{*}{L8} &  & - & - & - & - & - & Avg. \\
& & & & & & & & & & & & & Impr. \\ \hline
RR2+BW &13.7 & 12.8 & 12.4 & 12.2 & 12.2$\bullet$ & \textbf{12.2}$\dagger \bullet$ & - & - & - & - & - & - & -0.6\\ 
MLR+BW & 12.4$\bullet$ & 12.1 & \textbf{12.0}$\bullet$ & \textbf{11.9}$\bullet \star$ & 12.2$\bullet$ & \textbf{12.2}$\dagger \bullet$ & - & - & - & - & - & - & -1.1 \\ 
MLR+BW & \multirow{2}{*}{\textbf{12.3}} & \multirow{2}{*}{\textbf{11.9}$\bullet$} & \multirow{2}{*}{\textbf{12.0}$\bullet$} & \multirow{2}{*}{\textbf{11.9}$\bullet \star$} & \multirow{2}{*}{\textbf{12.1}$\bullet$} & \multirow{2}{*}{\textbf{12.2}$\dagger \bullet$} & & - & - & - & - & - & \multirow{2}{*}{-1.2} \\
+Untied & & & & & & & & & & & & & \\ \hline  \hline
\multirow{2}{*}{\textbf{CHiME-3}}  & \multirow{2}{*}{L3} & \multirow{2}{*}{L4} & \multirow{2}{*}{L5} & \multirow{2}{*}{L6} & \multirow{2}{*}{L7} & \multirow{2}{*}{L8} & \multirow{2}{*}{L9} & \multirow{2}{*}{L10} & \multirow{2}{*}{L11} & \multirow{2}{*}{L12} & \multirow{2}{*}{L13} & \multirow{2}{*}{L14} & Avg. \\
& & & & & & & & & & & & & Impr. \\ \hline
RR1+BW & 20.0 & 19.5 & \textbf{19.1} & \textbf{19.5} & \textbf{19.7} & \textbf{20.3} & \textbf{20.7} & 21.4 & 22.1 & 22.9 & 23.9 & 25.8$\dagger \bullet$ & -5.4 \\ 
MLR+BW &  19.8 & 19.5 & 19.5 & 19.7 & 20.2 & 20.4 & 20.9 & 21.5 & 22.2 & 23.4 & 24.9 & 25.7$\bullet$ & -5.2\\
MLR+BW & \multirow{2}{*}{\textbf{19.3}} & \multirow{2}{*}{\textbf{18.6}} & \multirow{2}{*}{19.2} & \multirow{2}{*}{\textbf{19.5}} & \multirow{2}{*}{20.0} & \multirow{2}{*}{\textbf{20.3}} & \multirow{2}{*}{20.8} & \multirow{2}{*}{\textbf{21.2}} & \multirow{2}{*}{\textbf{21.8}} & \multirow{2}{*}{\textbf{22.5}} & \multirow{2}{*}{\textbf{23.2}} & \multirow{2}{*}{\textbf{24.0}$\star$} & \multirow{2}{*}{-5.8} \\
+Untied  & & & & & & & & & & & & & \\ \hline  
\end{tabular}
\end{center}
\end{small}
\caption{\label{wertie} WER[\%] ($\downarrow$) results when the ties are broken using prior knowledge.  ``$\dagger$'' =  the result is not statistically different from random ROVER;  ``$\bullet$'' = the result is not statistically different from SysO; ``$\star$'' the result is not statistically different from SegO.}
\end{table*}

Similarly to what we observed in the previous experiments, the best results are obtained at low  levels of combination. This calls for solutions to automatically find (or at least approximate) the optimum level, which is 
the problem 
investigated in the next section.

\subsection{Optimum level of combination}
\label{subsec:optimumcomb}

Table~\ref{wertie} shows that, after breaking the ties, the optimum level of combination in IWSLT (with 8 components) is either $L4$ or $L6$, while in CHiME-3 (with 14 components) it is $L4$.  A \textit{post-hoc} comparison of the results achieved by each level on the test set, however, is only made possible by the availability of references to compute final WER scores. This observation raises a new practical issue: 
{how can we predict the optimum level of combination 
in a real condition in which we have no access to
the reference transcripts?} 

The problem of finding a dataset-specific stopping criterion, as an alternative to the simple and risky ``take-all'' strategy, is well motivated. 
Although
it seems less important for IWSLT, where  final WER scores are quite similar for all the levels, a method to avoid entering 
harmful 
inputs into the ROVER combination can significantly change the results in CHiME-3.

Several sub-optimal solutions can be applied to 
address this problem. The simplest one is the random choice, which
is acceptable in situations where hypothesis quality
is homogeneous (IWSLT) but,
as we will show below,
it can be 
inadequate when the variability is higher (CHiME-3). 
Another option is to 
determine the optimal level in the training set and apply it to the test data. This, however, would not be efficient when the number of components varies between training and test, as in the case of the IWSLT task.
A better solution
is suggested by
the findings of \cite{audhkhasi2014theoretical}, which demonstrates the strong dependency between the performance of  ASR system combination methods and the diversity and quality of the components. Following this intuition, 
we explore
a classification-based approach, in which a binary classifier is trained to learn whether a combination level is appropriate (labeled as \textit{1}) or not (labeled as \textit{0}). Based on the predictions of the classifier, for each given segment we select the appropriate level. In case of finding multiple optimum levels, we rely on the confidence score of the classifier.

In the following, we explain this classification task in detail, 
focusing on the data, features, classifiers and evaluation metric used in our final round of experiments. 

\paragraph{Data}
To prepare the training data, for each segment we first run the proposed segment-based QE-informed ROVER at all 
combination levels.
Then,  label \textit{1}  is assigned to the level(s) with the smallest WER. It is important to note that, for some segments, different levels may result in the same smallest WER score. When this happens,
multiple levels are labeled as \textit{1}
thus resulting in skewed label distributions. 
This is the case of IWSLT data, especially the test set, in which more levels have identical scores compared to CHiME-3 (the percentage of positive examples in the two test sets is respectively 77.0 and 57.3). 
Looking at the results in Table~\ref{wertie} this is not 
surprising, since 
the WER difference between the best  and the worst level is minimal (0.4\% for $MLR+BW+Untied$) compared to CHiME-3 (5.4\%). Moreover, looking at the  distributions in Figures~\ref{chanal11} and \ref{chanal12}, we notice that the majority of the IWSLT segments have transcriptions with small diversity.  These considerations 
indicate a higher probability to find transcriptions of similar quality in the IWSLT data and, in turn, 
a more skewed label distribution
when training our classifier. 

\paragraph{Features}
The features for this classification task are extracted by using the confusion networks generated by ROVER at each combination level. In particular, we compute:

\begin{enumerate}
\item  Overall diversity of the level, computed by using the method proposed in \cite{audhkhasi2014theoretical};
\item  Levenshtein distance between the first and the last hypothesis;
\item  Avg.  Levenshtein distance between the first hypothesis and the others;
\item  Avg.  Levenshtein distance between each hypothesis and the next one;
\item  Avg.  Levenshtein distance between each hypothesis and the final combination;
\item  Mean/min/max predicted WER obtained by RR methods among the hypotheses. 
\end{enumerate}

Except for the first feature, the others are quite simple to compute.
The approach described in \cite{audhkhasi2014theoretical} defines the {\it diversity} of the set of ASR systems to combine as  ``the average spread of the individual ASR predictions around the average prediction''.
Given a ROVER WTN (see \S \ref{rover}) formed by $I$ confusion slots and built by combining $M$ ASR systems, we define as ${\bf w}_i^m$ ($1\leq m\leq M, 1\leq i\leq I$) a one-hot binary vector, of dimension equal to the dictionary size,  having all $0$ components except for the one corresponding to the word furnished by system $m$ in the $i^{th}$ confusion slot.
First, the voting scheme of ROVER averages the ${\bf w}_i^m$ across the $M$ ASR systems, as follows:
\begin{equation}
\nonumber
{\bf w}_i^{avg}=\frac{1}{M}\sum_{m=1}^{M} {\bf w}_i^{m}
\end{equation}

\noindent
In this way  ${\bf w}_i^{avg}$ contains the frequencies of the words in the $i^{th}$ confusion slot. Then, as seen in  \S \ref{rover}, to provide the combined recognized string, ROVER outputs the word with the highest frequency in each slot $i$. 
The diversity measure of $M$ ASR hypotheses being combined is defined with the following Equation \cite{audhkhasi2014theoretical}. 
Note that, $\| \ \ \|_2^2$ represents the  vector $L2$ norm. 

{\begin{equation}
\label{eq:diver}
\nonumber
Diversity = \frac{1}{I\times M} \sum_{i=1}^{I}\sum_{m=1}^{M} \frac{1}{2} \left \| {\bf w}_i^{avg} - {\bf w}_i^m \right \|_2^2    
\end{equation}
}
\noindent

\paragraph{Classifiers}
We experiment with Support Vector Machine (SVM) \cite{WICS:WICS49}, Random Forest \cite{breiman2001random} and BayesNet \cite{friedman1997bayes} classifiers trained
 in binary mode.
 Parameters are tuned with 5-fold cross validation on the training set.

\paragraph{Evaluation metric}
Balanced accuracy, which is particularly appropriate in case of skewed distributions, is used as evaluation metric to optimize hyper-parameters and select the best  classification algorithm.
The cross validation results reported in Table~\ref{weropt}  show
that BayesNet outperforms the other classifiers and, by a large margin, also the 50.0\%  balanced accuracy reachable with a baseline system 
relying on majority voting. 
One possible explanation for the success of BayesNet could be in its higher capability to work with a limited number of features (8) and a large number of instances (4496 for IWSLT and 13476 for CHiME-3). Comparing the performance obtained on the two datasets, the fact that CHiME-3 results are better than the IWSLT ones suggests that the more skewed distribution of the IWSLT labels penalizes the classifier.

\begin{table*} [h]
\begin{small}
\begin{center}
\begin{tabular}{|l||c||c|}
\hline

\ \ \ \ \ \ \ \ \ \ \ \ \ \textbf{Task}& \textbf{IWSLT} &   \textbf{CHiME-3} \\ \hline
\textbf{Classifier}& Balanced Acc. IWSLT2012  & Balanced Acc. DT05 \\ \hline
SVM             & 52.8 & 66.1 \\
Random Forest   & 58.7 & 71.5  \\ 
BayesNet        & \textbf{65.5} & \textbf{72.0} \\ \hline

\end{tabular}
\end{center}
\end{small}
\caption{\label{weropt} Performance of different binary classifiers used to find the optimum level of combination.}
\end{table*}

In light of its higher balanced accuracy, the BayesNet classifier has been selected  to predict the best level of combination for each 
segment 
in the test data.
On both tasks the results outperform those of the sub-optimal strategy
based on  
random selection of the best combination level.
While on IWSLT the WER reduction is unsurprisingly minimal (from 12.1\% to 12.0\%), on CHiME-3 the gain is much larger (from 20.9\% to 18.8\%).\footnote{The baseline sub-optimal scores are obtained by averaging the results obtained from five iterations of the random selection process.}
More interestingly, besides outperforming 
random 
selection, our classifier provides predictions that closely approximate 
the performance of the best levels shown in Table~\ref{wertie}  (11.9\% for IWSLT and 18.6\% for CHiME-3).

Overall, these results indicate that:
\textit{i)} the optimal combination does not only depend on the quality of the hypotheses and the reliability of their ranking but also on their diversity (this is in line with the discussion in  \S \ref{chime3sec} and with the results reported in  \cite{audhkhasi2014theoretical});
\textit{ii)} the relative importance and contribution of these aspects can be learned from data, and 
\textit{iii)} the resulting models can effectively support our QE-informed ROVER approach in real operating conditions where reference transcripts are not available.

\section{Conclusions}
\label{conclusion}

We investigated the use of ASR quality estimation as a way to address 
some 
potential drawbacks of the current
 ROVER-based hypothesis combination strategies. These concern the influence of three main factors on the final result: \textit{i)} the availability of \textit{confidence information} about the inner workings of the ASR systems 
 producing the transcriptions, \textit{ii)} the \textit{order} in which the input hypotheses are used to initialise the fusion process and \textit{iii)} the \textit{granularity} of the combined transcriptions.
Our QE-based approach overcomes the first limitation by informing ROVER with automatic and confidence-independent quality predictions. This allows us to operate in ``black-box'' application scenarios in which the hypotheses to be combined come from unknown systems.
The reliability of our  predictions is exploited to overcome the second limitation by sorting the input hypotheses and prioritizing the most accurate ones to support ROVER's voting scheme. 
The third limitation is tackled 
by feeding ROVER with reliable rankings at segment-level instead of considering long utterance transcriptions whose quality can considerably vary at local level. This allows our QE-informed ROVER to better exploit the local diversity of the combined hypotheses.

By extending previous experiments to a new scenario, we analyzed 
the performance of different ranking strategies 
on two different tasks. 
Working on the automatic transcription of TED talks from the IWSLT evaluation campaign, we experimented with the combination of outputs from a \textit{single microphone and multiple systems}. Working with the automatic transcription of Wall Street Journal articles from the CHiME-3 campaign, we tested our 
capability to combine 
outputs from \textit{multiple microphones  and multiple systems}.

In all tasks, our 
approach  significantly outperforms
the standard random ROVER, which is fed with entire audio 
transcriptions presented in random order (absolute WER improvements range from 0.5\% to 7.3\%).
More interestingly, our best results 
are competitive with
those achieved by two strong oracles, which are  informed about the true ranks of the input hypotheses respectively at system  and segment level. 

To make our analysis more complete, 
we studied 
performance variations as a function of the diversity of the combined hypotheses and, finally, we addressed two practical issues: 
\textit{i)} dealing with tied ranks 
and \textit{ii)} automatically finding the optimum number of hypotheses that have to be combined for each input segment.
Our solutions respectively allowed us to isolate a unique, best performing ranking strategy 
and to 
deploy it for a competitive and fully automatic method to inform ROVER with QE predictions.

Our results open the door to new applications for ASR QE. Recently, we started to work on the unsupervised adaptation of deep neural networks  for acoustic modeling, using segment-level WER prediction  to  
regularize the objective function computed on the adaptation data.
ASR QE can also be used to select audio segments with {\em good} transcriptions to be inserted either in adaptation or in the training sets.
In an MDM scenario QE can be employed to select the best  microphone (similarly to what we did with CHiME-3 data) or 
to train improved acoustic models by filtering clean data with different predefined impulse responses.
In this context, QE-informed ROVER has been applied to improve the performance given by a more recent CHiME-3 baseline~\cite{hori2015,fala2016}.

\newpage

\bibliographystyle{model2-names}
\bibliography{sample2}

\end{document}